\documentclass[sigconf]{acmart}

\renewcommand\footnotetextcopyrightpermission[1]{} % removes footnote with conference information in first column
\usepackage{booktabs} % For formal tables
\usepackage{enumitem}
\usepackage{subcaption}
\usepackage[linesnumbered,lined,boxed,commentsnumbered]{algorithm2e}

\setcopyright{rightsretained}
\newcommand{\eg}{\textit{e.g., }}

\acmDOI{10.475/123_4}

\acmISBN{123-4567-24-567/08/06}

\acmPrice{15.00}
\begin{document}
\title{Call Attention to Rumors: Deep Attention Based Recurrent Neural Networks for Early Rumor Detection}
%\titlenote{Produces the permission block, and copyright information}
%\subtitle{Extended Abstract}
%\subtitlenote{The full version of the author's guide is available as \texttt{acmart.pdf} document}

\author{Tong Chen}
%\authornote{}
\orcid{}
\affiliation{%
  \institution{The University of Queensland}
  \streetaddress{}
  \city{Brisbane}
  \state{Queensland}
  \postcode{4072}
}
\email{tong.chen@uq.edu.au}

\author{Lin Wu}
\authornote{Corresponding author}
\affiliation{%
  \institution{The University of Queensland}
  \streetaddress{}
  \city{Brisbane}
  \state{Queensland}
  \postcode{4072}
}
\email{lin.wu@uq.edu.au}

\author{Xue Li}
%\authornote{-}
\affiliation{%
  \institution{The University of Queensland}
  \streetaddress{}
  \city{Brisbane}
  \state{Queensland}
  \postcode{4072}
}
\email{xue.li@itee.uq.edu.au}

\author{Jun Zhang}
%\authornote{-}
\affiliation{%
  \institution{Deakin University}
  \streetaddress{}
  \city{Melbourne}
  \state{Victoria}
  \postcode{3125}
}
\email{jun.zhang@deakin.edu.au}

\author{Hongzhi Yin}
%\authornote{-}
\affiliation{%
  \institution{The University of Queensland}
  \streetaddress{}
  \city{Brisbane}
  \state{Queensland}
  \postcode{4072}
}
\email{h.yin1@uq.edu.au}

\author{Yang Wang}
%\authornote{-}
\affiliation{%
  \institution{The University of New South Wales}
  \streetaddress{}
  \city{Sydney}
  \state{NSW}
  \postcode{2052}
}
\email{wangy@cse.unsw.edu.au}

%% The default list of authors is too long for headers}
%\renewcommand{\shortauthors}{T. Chen et al.}

\begin{abstract}
The proliferation of social media in communication and information dissemination has made it an ideal platform for spreading rumors. Automatically debunking rumors at their stage of diffusion is known as \textit{early rumor detection}, which refers to dealing with sequential posts regarding disputed factual claims with certain variations and highly textual duplication over time. Thus, identifying trending rumors demands an efficient yet flexible model that is able to capture long-range dependencies among postings and produce distinct representations for the accurate early detection. However, it is a challenging task to apply conventional classification algorithms to rumor detection in earliness since they rely on hand-crafted features which require intensive manual efforts in the case of large amount of posts. This paper presents a deep attention model on the basis of recurrent neural networks (RNN) to learn \textit{selectively} temporal hidden representations of sequential posts for identifying rumors. The proposed model delves soft-attention into the recurrence to simultaneously pool out distinct features with particular focus and produce hidden representations that capture contextual variations of relevant posts over time. Extensive experiments on real datasets collected from social media websites demonstrate that (1) the deep attention based RNN model outperforms state-of-the-arts that rely on hand-crafted features; (2) the introduction of soft attention mechanism can effectively distill relevant parts to rumors from original posts in advance; (3) the proposed method detects rumors more quickly and accurately than competitors.
\end{abstract}

%\begin{CCSXML}
%<ccs2012>
%  <concept_id>10010520.10010553.10010562</concept_id>
%  <concept_desc>Computer systems organization~Embedded systems</concept_desc>
%  <concept_significance>500</concept_significance>
%  <concept_id>10010520.10010575.10010755</concept_id>
%  <concept_desc>Computer systems organization~Redundancy</concept_desc>
%  <concept_significance>300</concept_significance>
%  <concept_id>10010520.10010553.10010554</concept_id>
%  <concept_desc>Computer systems organization~Robotics</concept_desc>
%  <concept_significance>100</concept_significance>
%  <concept_id>10003033.10003083.10003095</concept_id>
%  <concept_desc>Networks~Network reliability</concept_desc>
%  <concept_significance>100</concept_significance>
%</ccs2012>
%\end{CCSXML}
%
%\ccsdesc[500]{Computer systems organization~Embedded systems}

% We no longer use \terms command

\keywords{Early rumor detection, Recurrent neural networks, Deep attention models}

\maketitle

\section{Introduction}\label{sec:intro}

The explosive use of contemporary social media in communication has witnessed the widespread of rumors which can pose a threat to the cyber security and social stability.
For instance, on April 23rd 2013, a fake news claiming two explosions happened in the White House and Barack Obama got injured was posted by a hacked Twitter account named Associated Press. Although the White House and Associated Press assured the public minutes later the report was not true, the fast diffusion to millions of users had caused severe social panic, resulting in a loss of \$136.5 billion in the stock market\footnote{http://www.dailymail.co.uk/news/article-2313652/AP-Twitter-hackers-break-news-White-House-explosions-injured-Obama.html}. This incident of a false rumor showcases the vulnerability of social media on rumors, and highlights the practical value of automatically predicting the veracity of information.
%\footnote{https://en.wikipedia.org/wiki/Boston_Marathon_bombing}

\begin{figure}[hbt]
\begin{tabular}{c}
\includegraphics[height=1.5in, width=3in]{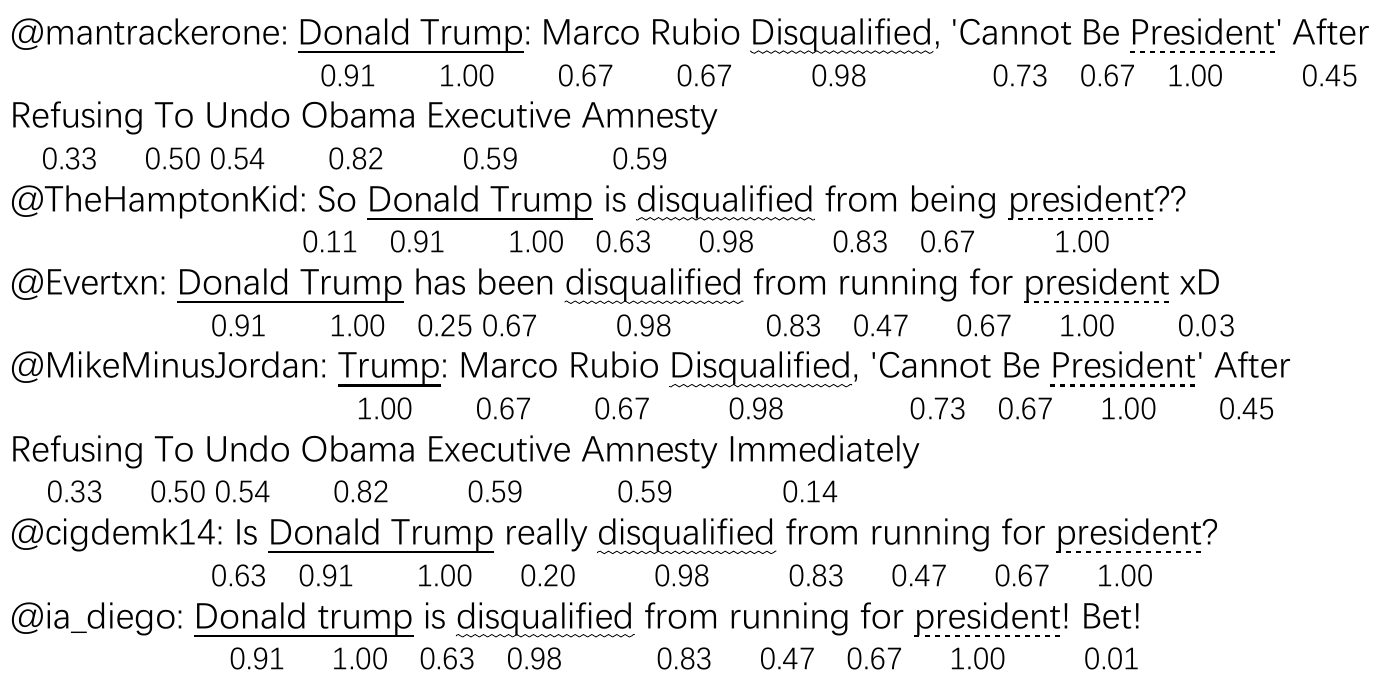}\\
(a) Streaming posts in regards to an event\\
\includegraphics[height=1in, width=3in]{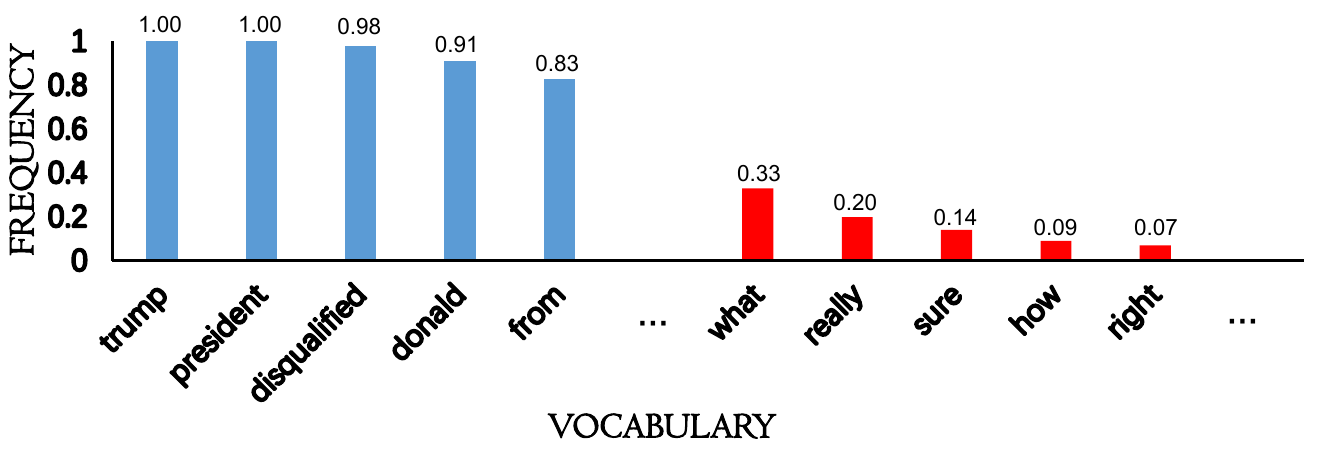}\\
(b) Statistics on textual phrases\\
\end{tabular}
\vspace{-0.4cm}
\caption{Posts from users on social media platforms exhibit duplication to great extent. For a specific event, \eg ``Trump being Disqualified from U.S. Election'', the texts of ``Donald Trump'', ``Obama'' and ``Disqualified'' appear very frequently in disputed postings.}\label{fig:duplicate}
\end{figure}

Debunking rumors at their formative stage is particularly crucial to minimizing their catastrophic effects.
Most existing rumor detection models employ learning algorithms that incorporate a wide variety of features and formulate rumor detection into a binary classification task.
They commonly craft features manually from the content, sentiment \cite{zimbra2016brand}, user profiles \cite{zafarani201510,YXLTIP17,YXLMM15}, and diffusion patterns of the posts \cite{wu2015false,ma2015detect,liu2015real,YXQCIKM13,YLXTNNLS17,kwon2013prominent}. Embedding social graphs into a classification model also helps distinguish malicious user comments from normal ones \cite{rayana2016collective,rayana2015collective}. These approaches aim at extracting distinctive features to describe rumors faithfully. However, feature engineering is extremely time-consuming, biased, and labor-intensive. Moreover, hand-crafted features are data-dependent, making them incapable of capturing contextual variations in different posts.

\begin{figure*}[hbt]
\centering
\includegraphics[height=2in, width=7in]{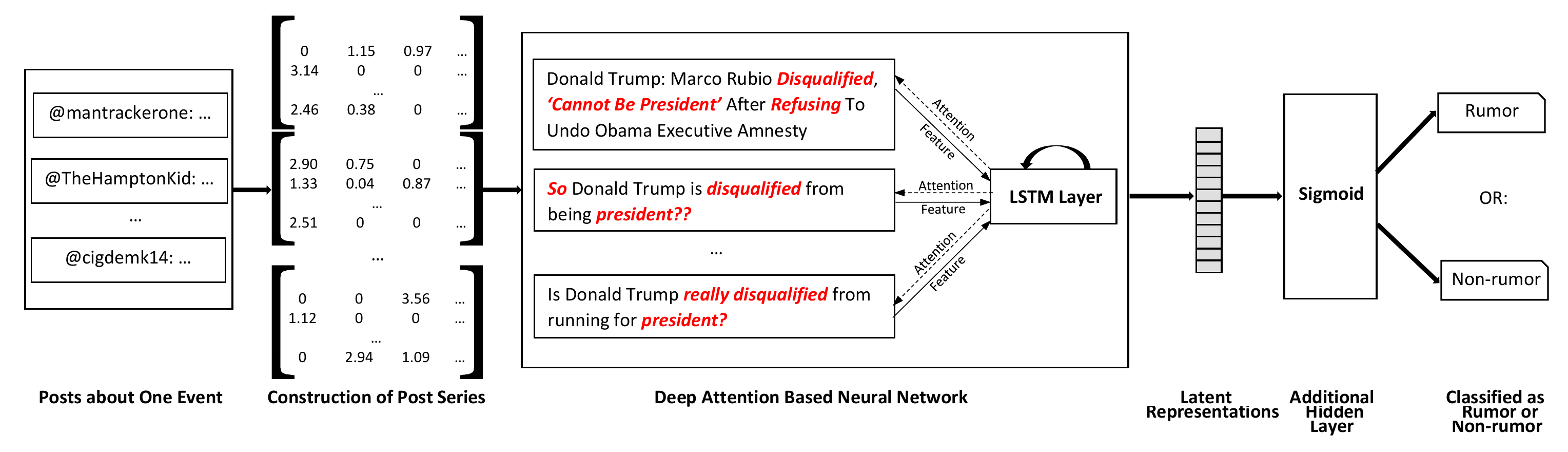}
\vspace{-0.9cm}
\caption{Schematic overview of our framework. In regards to each event, posts in sequence are collected and transformed into their tf-idf vectors. Then, deep recurrent neural networks augmented with soft-attention mechanism are deployed to derive temporal latent representations by capturing long-term dependency among post series and selectively focusing on important relevance. Additional layer is topped upon learned representations to determine the event to be rumor/non-rumor.}\label{fig:framework}
\vspace{-0.5cm}
\end{figure*}

More close examinations on rumors reveal that social posts related to an event under discussion are coming in the form of time series wherein users forward or comment on it continuously over time. As shown in Fig.\ref{fig:duplicate} (a), posts in regards to an event of US presidency are coming continuously along the event's timelines. Thus, to tackle with time series of posts, descriptive features should be extracted from contexts. However, as shown in Fig.\ref{fig:duplicate} (b), users' posts exhibit high duplication in their textual phrases due to the repeated forwarding, reviews, and/or inquiry behavior \cite{zhao2015enquiring}.  This poses a challenge on efficiently distilling distinct information from duplication, and flexible to capture their contextual variations as the rumor diffuses over time.

The propagation of information on social media has temporal characteristics, whilst most existing rumor detection methodologies ignore such a crucial property or are not able to capture the temporal dimension of data. One exception is \cite{ma2016detecting} where Ma \textit{et al.} uses an RNN to capture the dynamic temporal signals of rumor diffusion and learn textual representations under supervision. However, as the rumor diffusion evolves over time, users tend to comment differently in various stages, such as from expressing surprise to questioning, or from believing to debunking. As a consequence, textual features may change their importance with time and we need to determine which of them are more important to the detection task. On the other hand, the existence of duplication in textual phrases impedes the efficiency of training a deep network. Although some studies on duplication detection are available and effective in different tasks \cite{Duplicate-video,ImgCopy,Duplicate-photo}, these approaches are not applicable in our case where the duplication cannot be determined beforehand but is rather varied across post series over time. In this sense, two aspects of temporal long-term characteristic and dynamic duplication should be addressed simultaneously in an early rumor detection model.

\subsection{Challenges and Our Approach}

In summary, there are three challenges in early rumor detection to be addressed: (1) automatically learning representations for rumors instead of using labor-intensive hand-crafted features; (2) the difficulty of maintaining the long-range dependency among variable-length post series to build their internal representations; (3) the issue of high duplication compounded with varied contextual focus.
To combat these challenges, we propose a novel deep attention based recurrent neural network (RNN) for early detection on rumors, namely \textit{CallAtRumors} (\textbf{Call} \textbf{At}tention to \textbf{Rumors}). The overview of our framework is illustrated in Fig \ref{fig:framework}. Our model processes streaming textual sequences constructed by encoding contextual information from posts related to one event into a series of feature matrices. Then, the RNN with attention mechanism automatically learns latent representations by feed-forwarding each input weighted by attention probability distribution while adaptive to contextual variations. Moreover, an additional hidden layer with $sigmoid$ activation function using the learned latent representations predicts the event to be rumors or not.

Our framework is premised on the RNNs which are proved to be effective in recent machine learning tasks \cite{graves2013generating,Wake-sleep} in handling sequential data. This offers us the opportunity to automatically explore deep feature representations from original inputs for efficient rumor detection, thus avoiding the complexity of feature engineering. With attention mechanism, the proposed approach is able to selectively associate more importance with relevant features. Hence, we are able to tackle the problem in the context of high textual duplication and the efficiency of feature learning in early detection is ensured.

\subsection{Contributions}
The main contributions of our work are summarized as follows:
\begin{itemize}
\item We propose a deep attention based model that learns to perform rumor detection automatically in earliness. The model is based on RNN, and capable of learning continuous hidden representations by capturing long-range dependency an contextual variations of posting series.
\item The deterministic soft-attention mechanism is embedded into recurrence to enable distinct feature extraction from high duplication and advanced importance focus that varies over time.
\item We quantitatively validate the effectiveness of attention in terms of detection accuracy and earliness by comparing with state-of-the-arts on two real social media datasets: Twitter and Weibo.
\end{itemize}

The rest of the paper is organized as follows. Section \ref{sec:related} and Section \ref{sec:rnn} present the relationship to existing work and preliminary on RNN. We introduce the main intuition and formulate the problem  in Section \ref{sec:approach}. Section \ref{sec:exp} discusses the experiments and the results on effectiveness and earliness. We conclude this paper in Section \ref{sec:con} and points out future directions.

\section{Related Work}\label{sec:related}

Our work is closely connected with early rumor detection and attention mechanism. We will briefly introduce the two aspects in this section.

\subsection{Early Rumor Detection}
The problem of rumor detection  \cite{castillo2011information} can be cast as binary classification tasks. The extraction and selection of discriminative features significantly affects the performance of the classifier. Hu \textit{et al.} first conducted a study to analyze the sentiment differences between spammers and normal users and then presented an optimization formulation that incorporates sentiment information into a novel social spammer detection framework \cite{hu2014social}. Also the propagation patterns of rumors were developed by Wu \textit{et al.} through utilizing a message propagation tree where each node represents a text message to classify whether the root of the tree is a rumor or not \cite{wu2015false}. In \cite{ma2015detect}, a dynamic time series structure was proposed to capture the temporal features based on the time series context information generated in every rumor's life-cycle. However, these approaches requires daunting manual efforts in feature engineering and they are restricted by the data structure.

Early rumor detection is to detect viral rumors in their formative stages in order to take early action \cite{sampson2016leveraging}. In \cite{zhao2015enquiring}, some very rare but informative enquiry phrases play an important role in feature engineering when combined with clustering and a classifier on the clusters as they shorten the time for spotting rumors. Manually defined features has shown their importance in the research on real-time rumor debunking by Liu \textit{et al.} \cite{liu2015real}. By contrast, Wu \textit{et al.} proposed a sparse learning method to automatically select discriminative features as well as train the classifier for emerging rumors \cite{wugleaning}. As those methods neglect the temporal trait of social media data, a time-series based feature structure\cite{ma2015detect} is introduced to seize context variation over time. Recently, recurrent neural network was first introduced to rumor detection by Ma \textit{et al.} \cite{ma2016detecting}, utilizing sequential data to spontaneously capture temporal textual characteristics of rumor diffusion which helps detecting rumor earlier with accuracy. However, without abundant data with differentiable contents in the early stage of a rumor, the performance of these methods drops significantly because they fail to distinguish important patterns.

\subsection{Attention Mechanism}
As a rising technique in NLP (natural language processing) problems \cite{rocktaschel2015reasoning,yang2016hierarchical,sutskever2014sequence}, Bahdanau \textit{et al.} extended the basic encoder-decoder architecture of neural machine translation with attention mechanism to allow the model to automatically search for parts of a source sentence that are relevant to predicting a target word \cite{bahdanau2014neural}, achieving a comparable performance in the English-to-French translation task. Vinyals \textit{et al.} improved the attention model in \cite{bahdanau2014neural}, so their model computed an attention vector reflecting how much attention should be put over the input words and boosted the performance on large scale translation \cite{vinyals2015grammar}. In addition, Sharma \textit{et al.} applied a location softmax function \cite{sharma2015action} to the hidden states of the LSTM (Long Short-Term Memory) layer, thus recognizing more valuable elements in sequential inputs for action recognition. In conclusion, motivated by the successful applications of attention mechanism, we find that attention-based techniques can help better detect rumors with regards to both effectiveness and earliness because they are sensitive to distinctive textual features.
\vspace{0.4cm}
\section{Recurrent Neural Networks}\label{sec:rnn}
Recurrent neural networks, or RNNs \cite{rumelhart1988learning}, are a family of feed-forward neural networks for processing sequential data, such as a sequence of values
$x_1,...,x_\tau$. RNNs process an input sequence one element $x_t$ at a
time, updates the hidden units $h_t$, a ``state vector'' that implicitly contains information about the history of all the past elements of the sequence ($x_{<t}$), and generates output vector $o_t$ \cite{lecun2015deep}. The forward propagation begins with a specification of the initial state  $h_0$, then, for each time step $t$ from  $t = 1$ to $t = \tau$, the following update equations are applied \cite{Goodfellow-et-al-2016}:
\begin{equation}
\begin{split}
&h_t =\tanh(Ux_t + Wh_{t-1} + b), \\
&o_t = Vh_t+c,
\end{split}
\end{equation}
where parameters $U$, $V$ and $W$ are weight matrices for input-to-hidden, hidden-to-output and hidden-to-hidden connections, respectively. $b$ and  $c$ are the bias vectors. $\tanh(\cdot)$ is a hyperbolic tangent non-linear function.

The gradient computation of RNNs involves performing back-propagation through time (BPTT) \cite{rumelhart1988learning}. In practice, a standard RNN is difficult to be trained due to the well-known vanishing or exploding gradients caused by the incapability of RNN in capturing the long-distance temporal dependencies for the gradient based optimization \cite{DifficultRNN1994,LCAPR2017}. To tackle this training difficulty, an effective solution is to includes ``memory'' cells to store information over time, which are known as Long Short-Term Memory (LSTM) \cite{lstm1997,graves2013generating}. In this work, we employ LSTM as basic unit to capture long term temporal dependency among streaming variable-length post series.

\section{CallatRumors: Early Rumor Detection with Deep Attention based RNN}\label{sec:approach}

\begin{figure*}[hbt!]
%\centering
\begin{tabular}{ccc}
\hspace{-0.95cm}
\includegraphics[height=2in, width=2in]{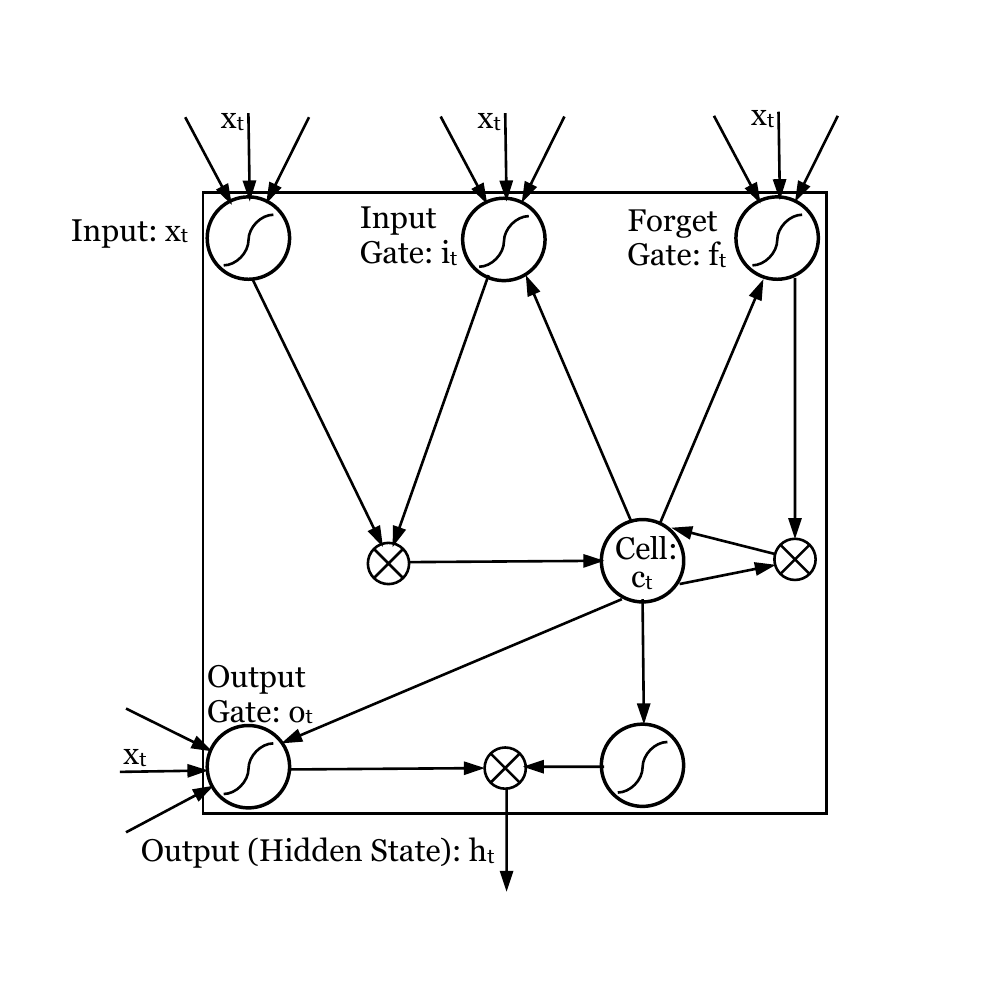}&
\hspace{-0.6cm}
\includegraphics[height=1.4in, width=2in]{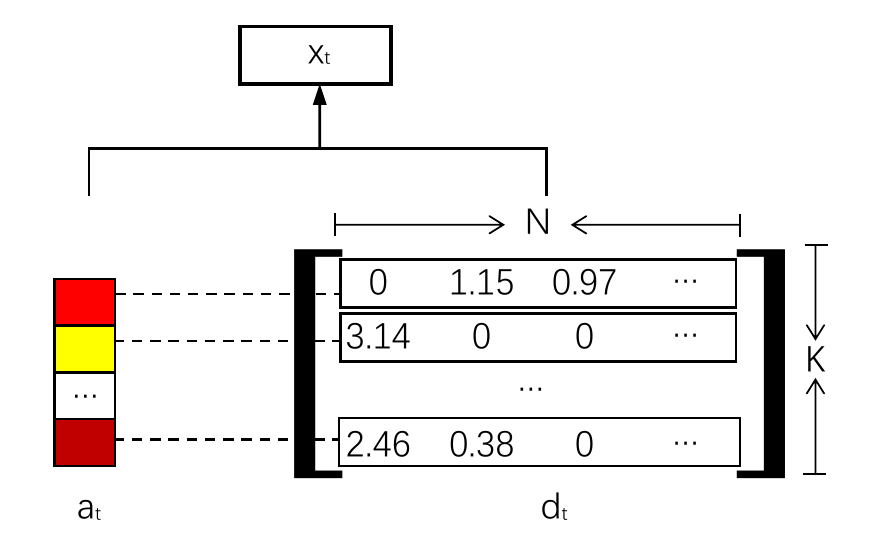}&
\hspace{-0.3cm}
\includegraphics[height=2in, width=3.5in]{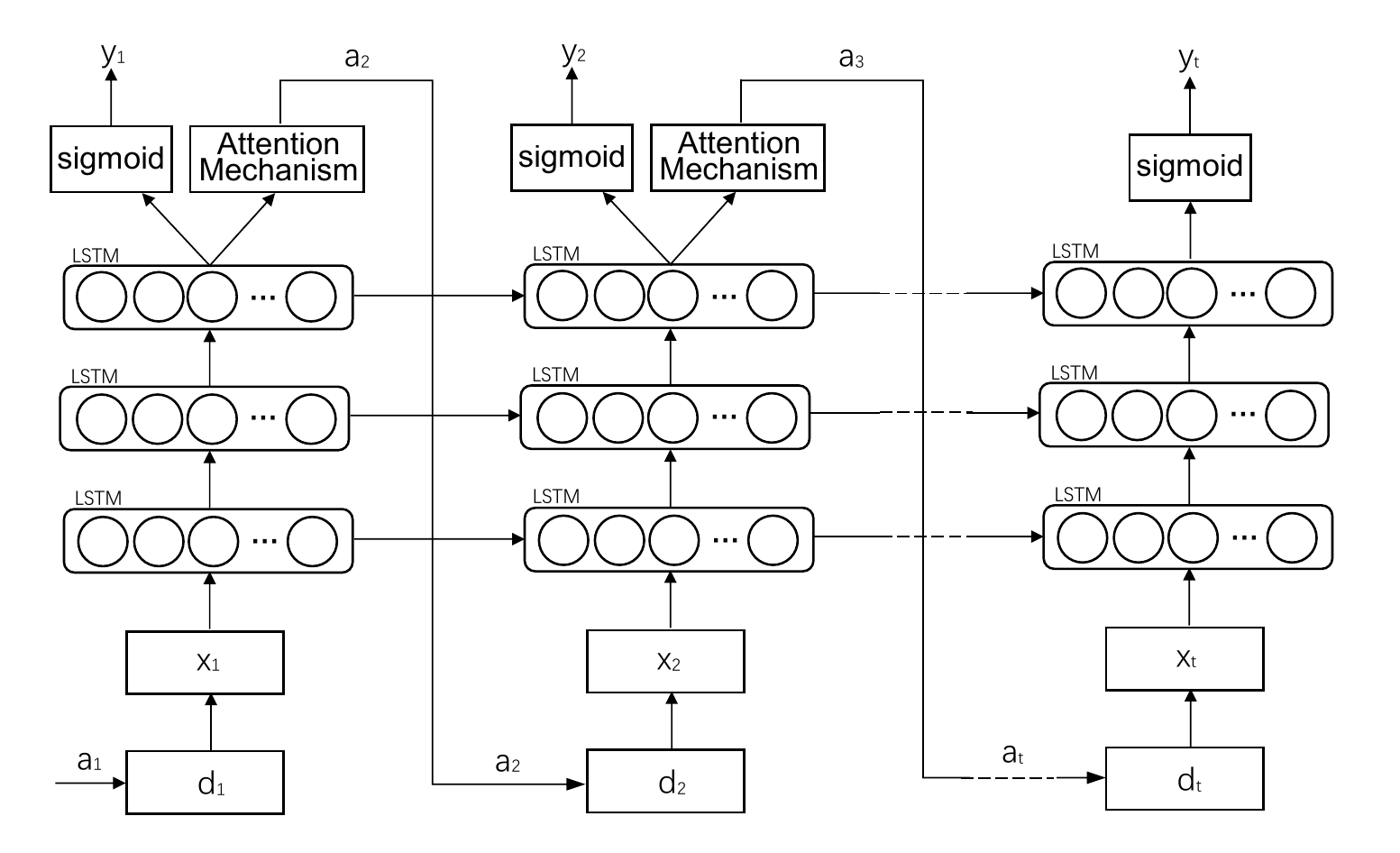}\\
(a) A LSTM unit & (b) The soft attention mechanism & (c) The proposed deep attention based recurrent model
\end{tabular}
\vspace{-0.3cm}
\caption{(a) A LSTM cell. Each cell learns how to weigh its input components (input gate), while learning how to modulate that contributions to the memory (input modulator). It also learns weights which erase the memory cell (forget gate), and weights which control how this memory should be emitted (output gate). (b) The attention module computes the current input $x_t$ as an average of the tf-idf features weighted according to the attention softmax $a_t$. (c) At each time stamp, the proposed model takes the feature slice $x_t$ as input and propagates $x_t$ through stacked layers of LSTM and predicts next location probability $a_{t+1}$ and class label $y_t$. }\label{fig:structure}
\vspace{-0.5cm}
\end{figure*}

In this section, we present the details of our framework with deep attention for classifying social textual events into rumors and non-rumors. First, we introduce a strategy that converts the incoming streams of social posts into continuous variable-length time series. Then, we describe the soft attention mechanism which can be embedded into recurrent neural networks to focus on selectively textual cues to learn distinct representations for rumor and/or non-rumor binary classification.

\subsection{Problem Statement}
Individual social posts contain very limited content due to their nature of shortness in context. On the other hand, a claim is generally associated with a number of posts that are relevant to the claim. These relevant posts regarding a claim can be easily collected to describe the central content more faithfully. Hence, we are interested in detecting rumor on an aggregate level instead of identifying each single posts \cite{ma2016detecting}. In other words, we focus on detecting rumors on event-level wherein sequential posts related to the same topics are batched together to constitute an event, and our model determines whether the event is a rumor or not.

Let $\boldsymbol E = \{E_i\}$ denote a set of given events, where each event $E_i=\{(p_{i,j},t_{i,j})\}_{j=1}^{n_i}$ consists of all relevant posts $p_{i,j}$ at time stamp $t_{i,j}$, and the task is to classify each event as  a rumor or not.

\subsection{Constructing Variable-Length Post Series}
For each event $E_i$, we collect a set of relevant post series to be the input of our model to learn latent representations. Within every event, posts are divided into time intervals, each of which is regarded as a batch. This is because it is not practical to deal with each post individually in the large number scale. To ensure a similar word density for each time step within one event, we group posts into batches according to a fixed post amount $N$ rather than slice the event time span evenly.

Algorithm \ref{algorithm:post_series} describes the construction of variable-length post series. Specifically, for every event $E_i=\{(p_{i,j},t_{i,j})\}_{j=1}^{n_i}$, post series are constructed with variable lengths due to different amount of posts relevant to different events. We set a minimum series length $Min$ to maintain the sequential property for all events. For events containing no less than ${N}{\times}{Min}$ posts, we iteratively take the first $N$ posts out of $E_i$ and feed them into a time interval $T$. The last ${n_i}-{N}{\lfloor}{\frac{n_i}{N}}{\rfloor}$ posts are treated as the last time interval. For events containing less than ${N}{\times}{Min}$ posts, we put ${\lfloor}{\frac{n_i}{Min}}{\rfloor}$ posts to the first $Min-1$ intervals and assign the rest into the last interval $T_{Min}$.

To model different words, we calculate the tf-idf (Term Frequency-Inverse Document Frequency) for the most frequent $K$ vocabularies within all posts. Proved to be an effective and lightweight textual feature, tf-idf is a numerical statistic that is intended to reflect how important a word is to a document in a collection or corpus \cite{leskovec2014mining}. In this case, for each post we have a $K$-word dictionary reflecting the importance of every word in a post, and the value is 0 if the word never appears in this post. Finally, every post is encoded by the corresponding K-word tf-idf dictionary, and within a specific internal a matrix of ${K}{\times}{N}$ can be constructed as the input of our model. If there are less than $N$ posts within an interval, we will expand it to the same scale by padding with $0$s. Hence, each set of post series consists of at least $Min$ feature matrices with a same size of $K$ (number of vocabularies) ${\times}$ ${N}$ (vocabulary feature dimension).

\IncMargin{1em}
\begin{algorithm}
\SetKwData{Left}{left}\SetKwData{This}{this}\SetKwData{Up}{up}
\SetKwFunction{Union}{Union}\SetKwFunction{FindCompress}{FindCompress}
\SetKwInOut{Input}{Input}\SetKwInOut{Output}{Output}

 \Input{Event-related posts $E_i=\{(p_{i,j},t_{i,j})\}_{j=1}^{n_i}$, post amount $N$, minimum series length $Min$}
 \Output{Post Series $S_i=\{{T_1} , ... , {T_v}\}$}
 /*Initialization*/\;
 $v = 1$; $x = 0$; $y = 0$\;
 \While{$true$}{
  \eIf{$n_i \geq N \times Min$}{
   \While{$v \leq {\lfloor}{\frac{n_i}{N}}{\rfloor}$}{
   $x = N \times (v-1) + 1$\;
   $y = N \times v$\;
   $Tv \leftarrow (p_{i,x} , ... , p_{i,y})$\;
   $v ++$\;
   }
   $Tv \leftarrow (p_{i,y+1} , ... , p_{i,n_i})$\;
   }{
   \While{$v < Min$}{
   $x = {\lfloor}{\frac{n_i}{Min}}{\rfloor} \times (v-1) + 1$\;
   $y = {\lfloor}{\frac{n_i}{Min}}{\rfloor} \times v$\;
   $Tv \leftarrow (p_{i,x} , ... , p_{i,y})$\;
   $v ++$\;
   }
   $Tv \leftarrow (p_{i,y+1} , ... , p_{i,n_i})$\;
  }
 }
 \Return $S_i$;
 \caption{Constructing Variable-Length Post Series}
 \label{algorithm:post_series}
\end{algorithm}
\DecMargin{1em}

\subsection{Long Short-Term Memory (LSTM) with Deterministic Soft Attention Mechanism}

To capture the long-distance temporal dependencies among continuous time post series, we employ Long Short-Term Memory (LSTM) unit \cite{graves2013generating,zaremba2014recurrent,xiang2017answer} to learn high-level discriminative representations for rumors. The structure of LSTM is formulated as
\begin{equation}\label{eq:lstm}
\begin{split}
& i_t = \sigma({U_i}{h_{t-1}} + {W_i}{x_t} + {V_i}{c_{t-1}} + b_i), \\
& f_t = \sigma({U_f}{h_{t-1}} + {W_f}{x_t} + {V_f}{c_{t-1}} + b_f), \\
& c_t = f_tc_{t-1} + i_t\tanh({U_c}{h_{t-1}} + {W_c}{x_t} + b_c), \\
& o_t = \sigma({U_o}{h_{t-1}} + {W_o}{x_t} + {V_o}{c_t} + b_o), \\
& h_t = o_t\tanh(c_t), \\
\end{split}
\end{equation}
where  $\sigma(\cdot)$ is the logistic sigmoid function, and
$i_t$, $f_t$, $o_t$, $c_t$ are the input gate, forget gate, output gate and cell input activation vector, respectively. In each of them, there are corresponding input-to-hidden, hidden-to-output, and hidden-to-hidden matrices: $U_{\bullet}$, $V_{\bullet}$, $W_{\bullet}$ and the bias vector $b_{\bullet}$. The LSTM architecture is essentially a memory cell which can maintain its state over time, and non-linear gating units can regulate the information flow into and out of the cell \cite{greff2016lstm}. A LSTM unit is shown graphically in Fig. \ref{fig:structure} (a).

In Eq.\eqref{eq:lstm}, the context vector $x_t$ is a dynamic representation of the relevant part of the social post input at time $t$. To calculate $x_t$, we introduce an attention weight $a_t[i],i=1,\ldots,K$, corresponding to the feature extracted at different element positions in a tf-idf matrix $d_t$. Specifically, at each time stamp $t$, our model predicts $a_{t+1}$, a softmax over $K$ positions, and $y_t$, a softmax over the binary class of rumors and non-rumors with an additional hidden layer with $sigmoid(\cdot)$ activations (see Fig.\ref{fig:structure} (c)). The location softmax \cite{sharma2015action} is thus, applied over the hidden states of the last LSTM layer to calculate $a_{t+1}$, the attention weight for the next input matrix $d_{t+1}$:
\begin{equation}\label{eq:location_softmax}
a_{t+1}[i] = P(L_{t+1}=i|h_t) = \frac{e^{{W_i}^\top{h_t}}}{\sum_{j=1}^{K}e^{W_j^{\top}h_t}} \qquad i \in 1,...,K,
\end{equation}
where $a_{t+1}[i]$ is the attention probability for the $i$-th element (word index) at time step $t+1$, $W_i$ is the weight matrix allocated to the $i$-th element, and $L_{t+1}$ is a random variable which represents the word index and takes 1-of-K values.
The attention vector $a_{t+1}$ is a probability distribution, representing the importance attached to each word in the input matrix $d_{t+1}$. Our model is optimized to assign higher focus to words that are believed to be distinct in learning rumor/non-rumor representations. After calculating these probabilities, the \textbf{soft deterministic attention} mechanism \cite{bahdanau2014neural} computes the expected value of the input at the next time step $x_{t+1}$ by taking expectation over the word matrix at different positions:
\begin{equation}\label{eq:soft_attention}
x_{t+1} = \mathbb{E}_{P(L_{t+1}|h_t)}[d_{t+1}] = \sum_{i=1}^{K}a_{t+1}[i] d_{t+1}[i],
\end{equation}
where $d_{t+1}$ is the input matrix at time step $t+1$ and $d_{t+1}[i]$ is the feature vector of the $i$-th position in the matrix $d_{t+1}$. Thus, Eq.\eqref{eq:soft_attention} formulates a deterministic attention model by computing a soft attention weighted word vector $\sum_i a_{t+1}[i] d_{t+1}[i]$. This corresponds to feeding a soft-$a$-weighted context into the system, whilst the whole model is smooth and differential under the deterministic attention, and thus learning end-to-end is trivial by using standard back-propagation.

We remark that attention models can be classified into soft attention and hard attention models. Soft attention models are shown to be deterministic and can be trained by using back-propagation whereas hard attention models are stochastic and the training requires the REINFORCE algorithm \cite{MnihNIPS2014} or by maximizing a variational lower bound or using importance sampling \cite{Wake-sleep,ShowAttendTell}. If we use hard attention models, we should sample $L_t$ from a softmax distribution of Eq.\eqref{eq:location_softmax}. The input $x_{t+1}$ would then be the feature at the sampled location instead of taking expectation over all the elements in $d_{t+1}$. Apparently, hard attention solutions are not differentiable and have to resort to some sampling, and thus we deploy soft attention in our model.

\subsection{Loss Function and Model Training}

In model training, we employ cross-entropy loss coupled with the doubly stochastic regularization \cite{ShowAttendTell} that encourages the model to pay attention to every element of the input word matrix. This is to impose an additional constraint over the location softmax, so that $\sum_{t=1}^{\tau}{a_{t,i}}\approx 1$. The loss function is defined as follows:
\begin{equation}\label{eq:loss}
\mathcal{L}=-\sum_{t=1}^{\tau}\sum_{i=1}^C y_{t,i} \log \hat{y}_{t,i} + \lambda \sum_{i=1}^K (1-\sum_{t=1}^{\tau} a_{t,i})^2 + \gamma  \phi^2,
\end{equation}
where $y_t$ is the one hot label vector, $\hat{y}_t$ is the vector of binary class probabilities at time stamp $t$, $\tau$ is the total number of time stamps, $C=2$ is the number of output classes (rumors or non-rumors), $\lambda$ is the attention penalty coefficient, $\gamma$ is the weight decay coefficient, and $\phi$ represents all the model parameters.

The cell state and the hidden state for LSTM are initialized using the input tf-idf matrices for faster convergence:
\begin{equation}\label{eq:initialization}
\begin{split}
c_0=f_c \left(\frac{1}{\tau}\sum_{t=1}^{\tau} \left(\frac{1}{K} \sum_{i=1}^K d_t[i]\right)\right), \\
h_0=f_h \left(\frac{1}{\tau}\sum_{t=1}^{\tau} \left(\frac{1}{K} \sum_{i=1}^K d_t[i]\right)\right),
\end{split}
\end{equation}
where $f_c$ and $f_h$ are two multi-layer perceptrons, and $\tau$ is the number of time stamps in the model. These values are used to compute the first location softmax $a_1$ which determines the initial input $x_1$.

\section{Experiments}\label{sec:exp}
This section reports how we evaluate the performance of our proposed methodology using real-world data collected from two different social media platforms. We first describe the construction datasets, and then perform self-evaluation to determine optimal parameters. Finally, we assess the effectiveness and efficiency of our model, CallAtRumors, by comparing with state-of-the-art methods.

\subsection{Datasets}

\begin{figure*}[!hbt]
\center
\includegraphics[trim={0cm 0cm 0cm 1cm}, height=1.7in, width=7in]{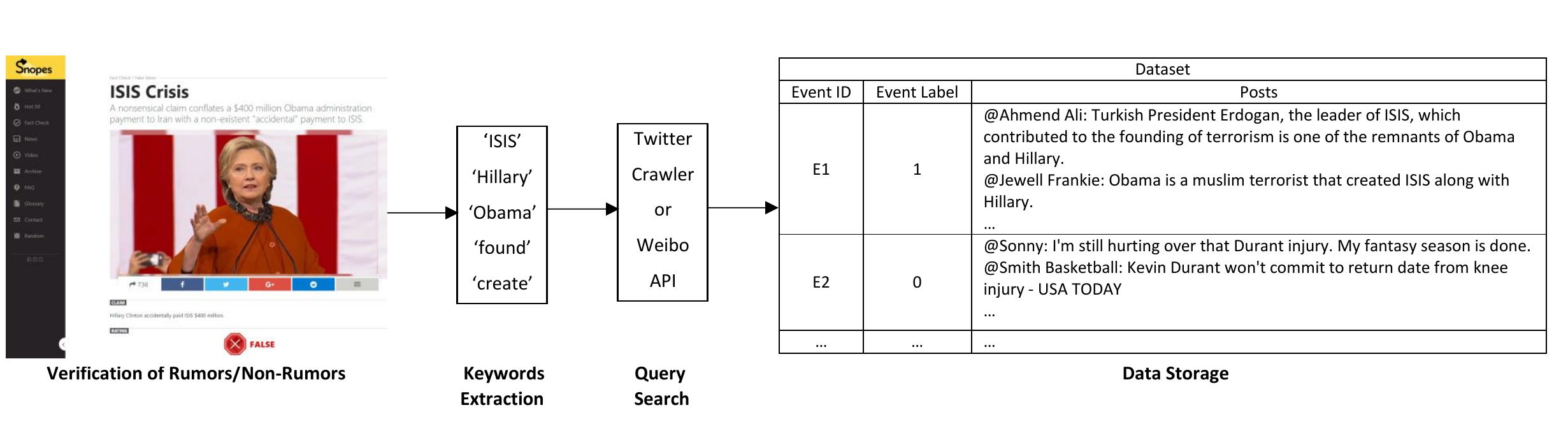}
\vspace{-1cm}
\caption{Data collection and dataset structure. For each event, its authenticity is verified through official news verification services. Then we manually extract suitable keywords for each event to ensure a precise search result of relevant posts. After that, we crawl posts with query search and store our collected data using the data storage structure shown in the table. Rumor events are labelled as 1 and normal events are labelled as 0.}
\vspace{-0.4cm}
\label{Figure:Data_Collection}
\end{figure*}

We use two public datasets published by \cite{ma2016detecting}. The datasets are collected from Twitter\footnote{www.twitter.com} and Sina Weibo\footnote{www.weibo.com} respectively. Both of the datasets are organised at event-level, in which the posts related to the same events are aggregated, and each event is labeled to 1 for rumor and 0 for non-rumor. In the following, we describe how the two datasets are originally constructed and how we expand them:
\begin{itemize}
\item  In the Twitter dataset, 498 rumors are collected using the keywords extracted from verified fake news published on Snopes\footnote{www.snopes.com}, a real-time rumor debunking website. It also contains 494 normal events from Snopes and two public datasets \cite{castillo2011information,kwon2013prominent}. For each event, the keywords are extracted and manually refined until the composed queries can have precise Twitter search results \cite{ma2016detecting}. All labelled events and related Tweet IDs are published by the authors, however some Tweets are no longer available when we crawled those Tweets, causing a 10\% shrink on the scale of data compared with the original Twitter dataset.
\item The Weibo dataset contains 2,313 rumors and 2,351 non-rumors. The polarity of all events are verified on Sina Community Management Center\footnote{http://service.account.weibo.com}. Then the keywords are manually summarized and modified for comprehensive post search for data collection using Weibo API.
\end{itemize}

In addition, to balance the ration of rumors and non-romors, we follow the criteria from \cite{ma2016detecting} to manually gather 4 non-rumors from Twitter and 38 rumors from Weibo to achieve a 1:1 ratio of rumors to non-rumors. The data collection procedure and our dataset structure are shown in Figure ~\ref{Figure:Data_Collection}\footnote{The webpage in this figure is downloaded from http://www.snopes.com/hillary-clinton-accidentally-gave-isis-400-million/}.

\begin{table}[t!]
  \begin{tabular}{|c|c|c|}
    \hline
    Statistic & Twitter & Weibo\\
    \hline
    Involved Users & 466,577 & 2,755,491\\
    Total Posts & 1,046,886 & 3,814,329\\
    Total Events & 996 & 4,702\\
    Total Rumors & 498 & 2,351\\
    Total Non-Rumors & 498 & 2,351\\
    Average Posts per Event & 1,051 & 811\\
    Minimum Posts per Event & 8 & 10\\
    Maximum Posts per Event & 44,316 & 59,318\\
    \hline
\end{tabular}
\caption{Statistical details of datasets}
\label{table:Datasets}
\vspace{-1.05cm}
\end{table}

Table ~\ref{table:Datasets} gives statistical details of the two datasets. We observe that more than 80\% of the users tend to repost the original news with very short comments to reflect their attitudes towards those news. As a consequence, the contents of the posts related to one event are mostly duplicate, triggering scarcity when extracting distinctive textual patterns within overlapping context. However, by implementing textual attention mechanism, CallAtRumors is able to lay more emphasis on discriminative words, and can guarantee high performance in such case.

\subsection{Self Evaluations}

\begin{figure}
\includegraphics[trim={0.5cm 1cm 0.5cm 1cm},clip, height=1.5in, width=1.88in]{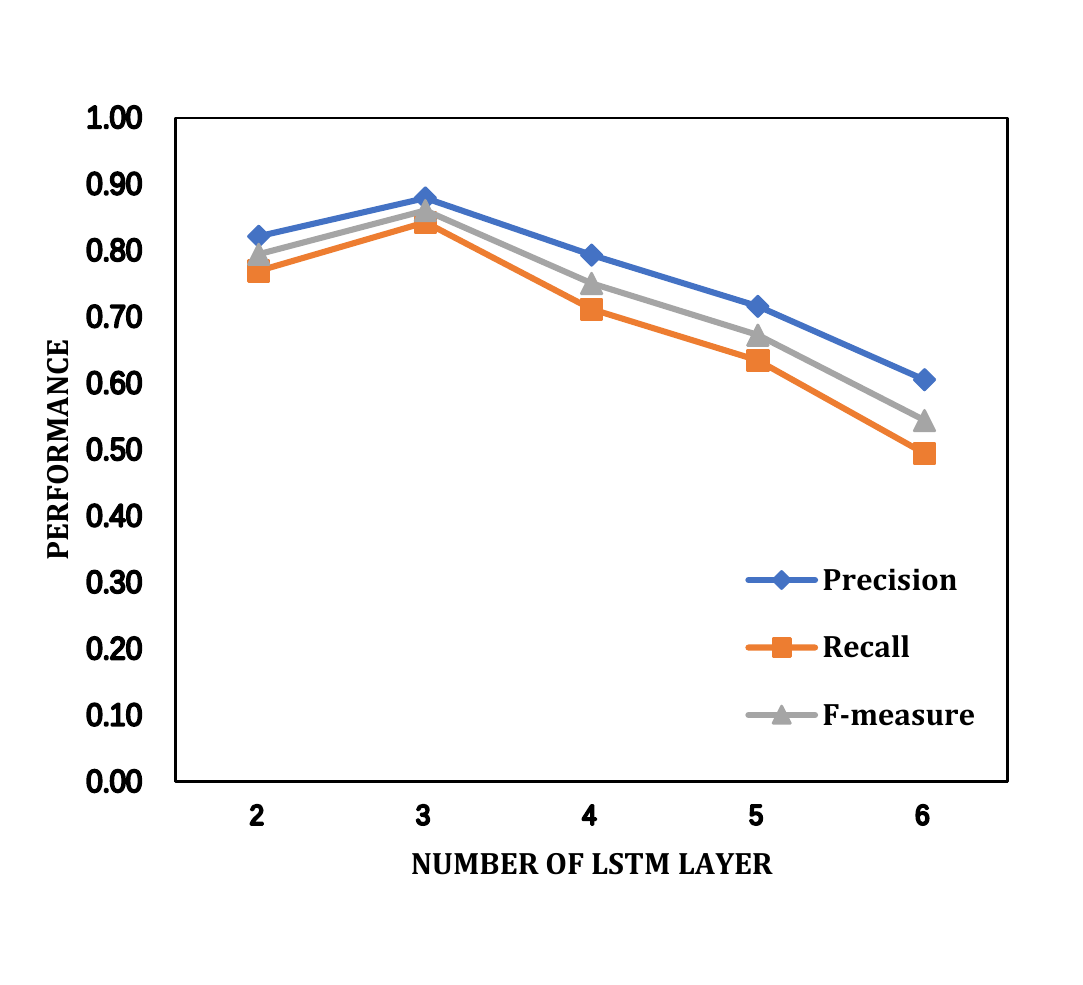}
\vspace{-0.4cm}
\caption{Results w.r.t varied number of LSTM layers. The best result can be achieved in the case of a three-layer LSTM model with1,024, 512 and 64 hidden states, respectively.}
\label{figure:LSTM-layers}
\end{figure}

The model is implemented by using Theano\footnote{http://deeplearning.net/software/theano/}. All parameters are set using cross-validation. To generate the input variable-length post series, we set the amount of posts $N$ for each time step as 50 and the minimum post series length $Min$ as 5. We selected $K$=10,000 top words for the construction tf-idf matrices. Apart from lowercasing, we do not apply any other special preprocessing like stemming \cite{bahdanau2014neural}. The recurrent neural network with attention mechanism can automatically learn to ignore those unimportant or irrelevant expressions in the training procedure.

For a hold-out dataset occupying 15\% of the events in each dataset, a self evaluation is performed to optimize the number of LSTM layers by varying the number of layers from 2 to 6. Results are shown in Figure \ref{figure:LSTM-layers}. Thus, we apply a three-layer LSTM model with descending numbers of hidden states of 1024, 512 and 64 respectively. The learning rate is set as 0.45 and we apply a dropout \cite{srivastava2014dropout} of 0.3 at all non-recurrent connections. For attention penalty coefficient, we set $\lambda$ to be 1.5, and the weight decay is set to be $10^{-5}$. Our model is trained by measuring the derivative of the loss through back-propagation \cite{collobert2011natural} algorithm, namely the Adam optimization algorithm \cite{kingma2014adam}. We iterate the whole training procedure until the loss value converges.

\begin{figure*}[!tbh]
\centering
\includegraphics[trim={0cm 0cm 0cm 1cm},height=1.9in, width=6.7in]{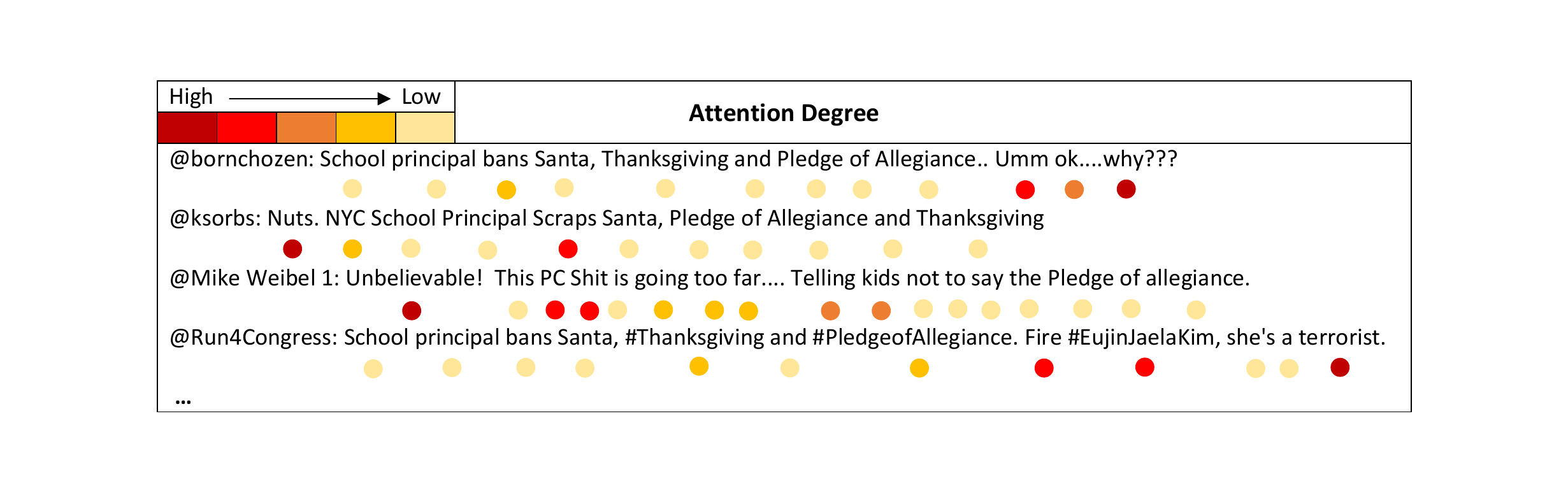}
\vspace{-1.3cm}
\caption{Visualization on varied attention on a detected rumor. Different color degrees reflect various attention degrees paid to each word in a post. In the rumor ``School Principal Eujin Jaela Kim banned the Pledge of Allegiance, Santa and Thanksgiving'', most of the vocabularies closely connected with the event itself are given less attention weight than words expressing users' doubting, esquiring and anger caused by the rumor. Our model learns to focus on expressions more useful in rumor detection while ignore unrelated words.}
\label{figure:Visualization}
\vspace{-0.5cm}
\end{figure*}

\subsection{Settings and Baselines}
We evaluate the effectiveness and efficiency of CallAtRumors by comparing with the following state-of-the-art approaches in terms of precision, recall and F-measure.

\begin{itemize}
\item DT-Rank \cite{zhao2015enquiring}: This is a decision-tree based ranking model, and is able to identify trending rumors by recasting the problem as finding entire clusters of posts whose topic is a disputed factual claim. We implement their enquiry phrases and features to make it comparable to our method.

\item SVM-TS \cite{ma2015detect}: This is a SVM (support vector machine) model that uses time-series structures to capture the variation of social context features. SVM-TS can capture the temporal characteristics of these features based on the time series of rumors' lifecycle with time series modelling technique applied to incorporate carious social context information. We use the features provided by them from contents, users and propagation patterns.

\item LK-RBF \cite{sampson2016leveraging}: To tackle the problem of implicit data without explicit links and jointed conversations, Sampson \textit{et al.} proposed two methods based on hashtags and web links to aggregate individual tweets with similar keywords from different threads into a conversation. We choose the link-based approach and combine it with the RBF (Radial Basis Function) kernel as a supervised classifier because it achieved the best performance in their experiments.

\item ML-GRU \cite{ma2016detecting}: This method utilizes recurrent neural networks to automatically discover deep data representations for efficient rumor detection. It also allows for early rumor detection with efficiency. Following the settings in their work, we choose the multi-layer GRU (gated recurrent unit) as baseline which shows the best result in the effectiveness and earliness test.

\item CERT \cite{wugleaning}: This is a cross-topic emerging rumor detection model which can jointly cluster data, select features and train classifiers by using the abundant labeled data from prior rumors to facilitate the detection of an emerging rumor. CERT is capable of extracting useful patterns in the case of data scarcity. Since CERT requires Tweet instances instead of event-level data, we use the tf-idf feature vector of the all the Tweets in one event to construct the feature matrix as required.
\end{itemize}

We hold out 15\% of the events in each dataset for cross-validation, and split the rest of  data with a ratio of 3:2 for training and test respectively. In particular, we keep the ratio between rumor events and normal events in both training and test set as 1:1. In the test on the effectiveness of CallAtRumors, all posts within each event are used during training and evaluation. In the study of efficiency, we take different ratios of the posts starting from the first post within all events, ranging from 10\% to 80\% in order to test how early CallAtRumors can detect rumors successfully.

\begin{figure*}[!tbh]
\begin{tabular}{cccc}
 \hspace{-1cm}\includegraphics[height= 1.5in, width= 1.88in]{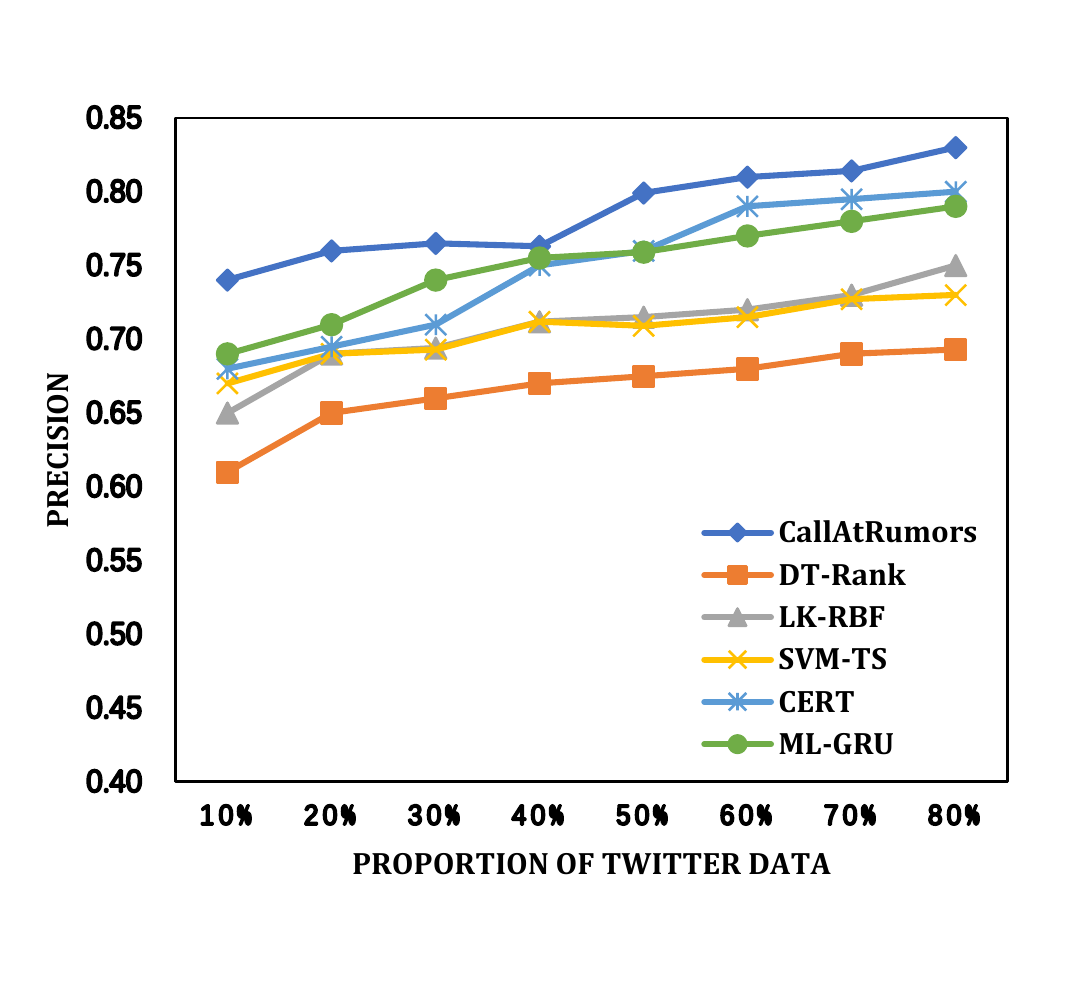}&
 \hspace{-0.5cm}
 \includegraphics[height= 1.5in, width= 1.88in]{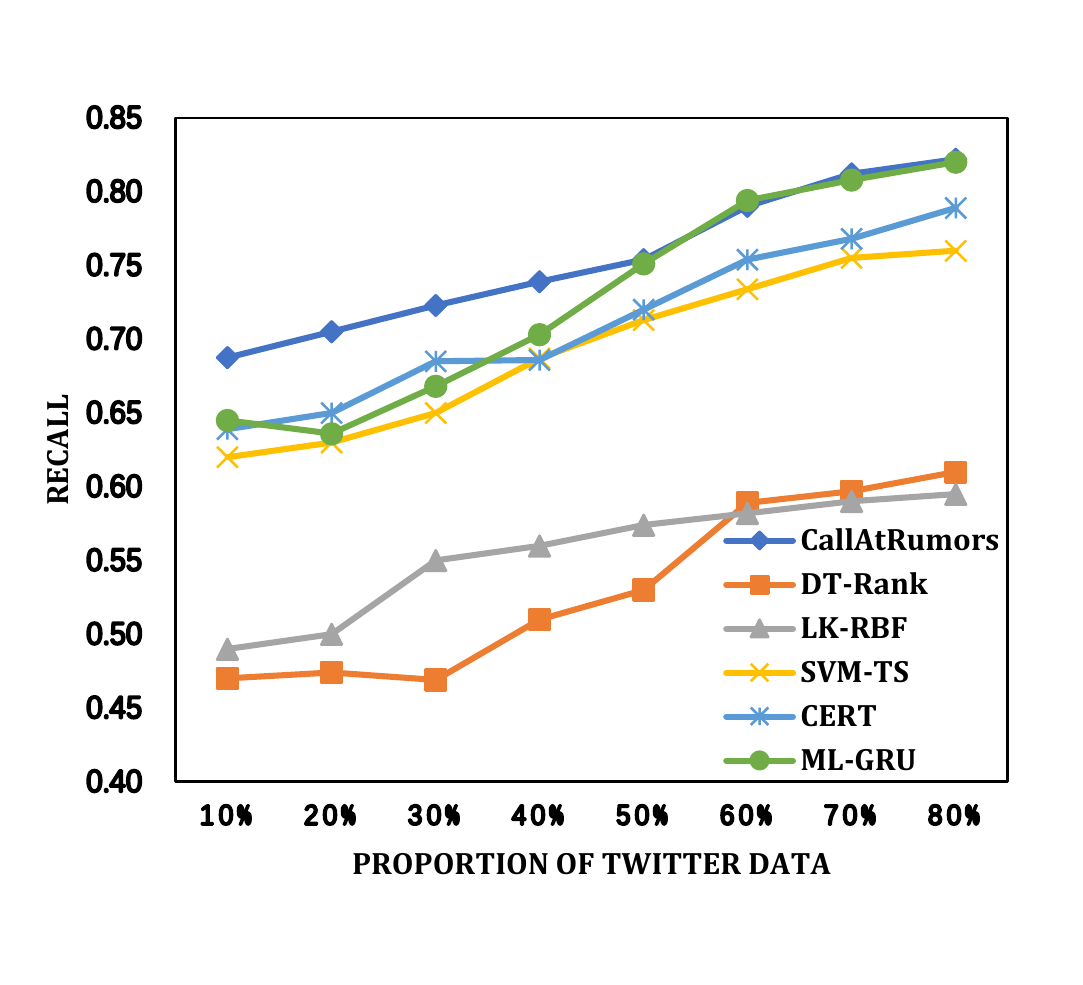}&
\hspace{-0.5cm}
 \includegraphics[height= 1.5in, width= 1.88in]{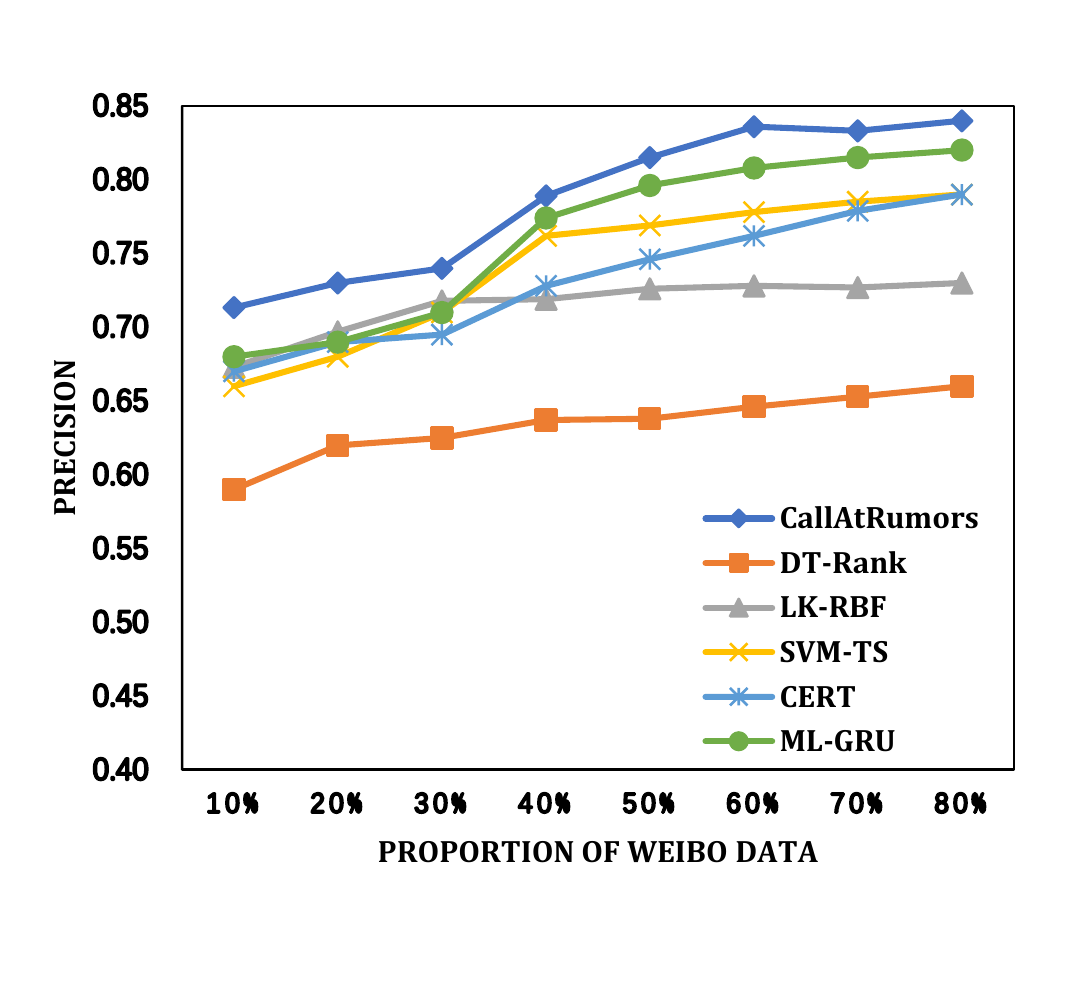}&
\hspace{-0.5cm}
 \includegraphics[height= 1.5in, width= 1.88in]{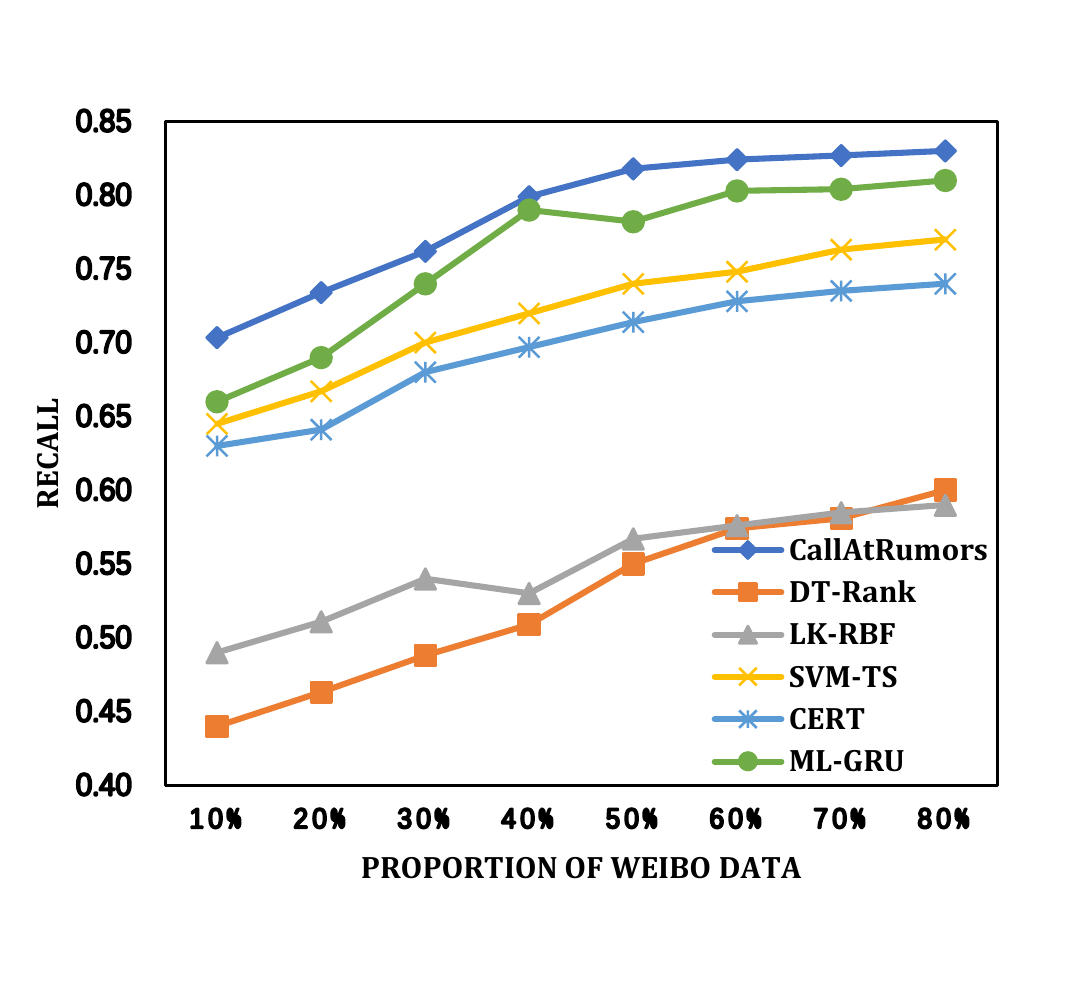}  \vspace{-0.5cm}\\
 (a) Precision on partial &(b) Recall on partial &(c) Precision on partial &(d) Recall on partial \\
 Twitter data & Twitter data & Weibo data & Weibo data
 \end{tabular}
\caption{Results of early rumor detection.}\label{figure:results}
\end{figure*}

\subsection{Effectiveness Validation}

\begin{table}
  \begin{tabular}{|c|c|c|c|}
    \hline
    Method & Precision & Recall & F-measure\\
    \hline
    DT-Rank & 71.50\% & 63.41\% & 0.6721\\
    LK-RBF & 78.54\% & 60.52\% & 0.6836\\
    SVM-TS & 76.33\% & 77.92\% & 0.7712\\
    CERT & 81.12\% & 79.66\% & 0.8038\\
    ML-GRU & 80.87\% & 82.97\% & 0.8191\\
    CallAtRumors & 88.63\% & 85.71\% & 0.8715\\
    \hline
\end{tabular}
\caption{Performance on the Twitter dataset}
\label{table:Performance on Twitter Dataset}
\end{table}

\begin{table}
  \begin{tabular}{|c|c|c|c|}
    \hline
    Method & Precision & Recall & F-measure\\
    \hline
    DT-Rank & 67.24\% & 61.33\% & 0.6415\\
    LK-RBF & 75.49\% & 61.08\% & 0.6752\\
    SVM-TS & 80.69\% & 78.26\% & 0.7946\\
    CERT & 79.70\% & 76.32\% & 0.7797\\
    ML-GRU & 82.44\% & 81.58\% & 0.8301\\
    CallAtRumors & 87.10\% & 86.34\% & 0.8672\\
    \hline
\end{tabular}
\caption{Performance on the Weibo dataset}
\label{table:Performance on Weibo Dataset}
\end{table}

Table ~\ref{table:Performance on Twitter Dataset} and Table ~\ref{table:Performance on Weibo Dataset} shows the performance of all approaches on Twitter dataset and Weibo dataset respectively. DT-Rank cannot effectively distinguish rumor from normal events when facing datasets with duplication in contents and scarcity in textual features. LK-RBF and SVM-TS achieve better results, indicating the ability of feature engineering to help classifiers detect rumors better. However, both LK-RBF and SVM-TS show the lack of adequate recall which represents how sensitive the models are towards rumors. Since CERT can jointly select discriminative features and train the topic-independent classifier with selected features \cite{wugleaning}, it achieves better results than the former three approaches in our datasets. The ML-GRU is competitive in both precision and recall due to its capability of processing sequential data and learning hidden states from raw inputs.

On the Twitter dataset, CallAtRumors outperforms competitors by achieving the precision, recall and F-measure of 88.63\%, 85.71\% and 0.8694 respectively. The same result can be seen on the Weibo dataset, where CallAtRumors achieves the precision, recall and F-measure of 87.10\%, 86.34\% and 0.8672 respectively. Figure ~\ref{figure:Visualization} illustrates the intermediate attention results on different words within a detected rumor event. The effectiveness validation proves the affect of attention mechanism in making LSTM units sensitive to distinctive words and tokens by associating more importance to certain locations in every feature matrix against other ones.

\subsection{More Comparison with the State-of-the-art: CERT \cite{wugleaning}}
To demonstrate how the conditions of datasets affect the performance of rumor detection, we compare the performance of CallAtRumors with CERT using different datasets. To reproduce the same experimental conditions as \cite{wugleaning}, we have also organized a sample dataset using the criteria described in the work. We use queries generated from 220 reported rumors on Snopes and regular expressions belonging to the same topics to crawl 7,580 Tweets and manually labeled each Tweet by reading the content and referring to the Snopes article. At last we result in a sample dataset containing 1,193 rumor Tweets and 6,387 non-rumor Tweets, which also has a similar ratio of rumors to non-rumors as the dataset in \cite{wugleaning}.
Table ~\ref{table:compare_cert} shows the different results when CallAtRumors and CERT are applied to this sample dataset and our Twitter dataset. The results further explains our model's capability of capturing valuable patterns within our large-scale duplicated datasets by applying attention to more representative words.

\begin{table}
  \begin{tabular}{|c|c|c|c|c|}
    \hline
    Method & Dataset & Precision & Recall & F-measure\\
    \hline
    CallAtRumors & Sample & 91.82\% & 89.43\% & 0.9061\\
    CERT & Sample & 90.35\% & 85.78\% & 0.8801\\
    CallAtRumors & Twitter & 88.63\% & 85.71\% & 0.8715\\
    CERT & Twitter & 81.12\% & 79.66\% & 0.8038\\
    \hline
\end{tabular}
\caption{More comparison with CERT \cite{wugleaning} on the Sample and Twitter datasets}\label{table:compare_cert}
\vspace{-1.3cm}
\end{table}

\subsection{Earliness Analysis}

In this experiment, we study the property of our approach in its earliness. To have fair comparison, we allow exiting rumor detection methods to be trained on rumors that are for evaluation. Through incrementally adding training data in the chronological order, we are able to estimate the time that our method can detect emerging rumors. The results on earliness are shown in Fig ~\ref{figure:results}.
At the early stage with 10\% to 60\% training data, CallAtRomors outperforms four comparative methods by a noticeable margin. In particular, compared with the most relevant method of ML-GRU, as the data proportion ranging from 10\% to 20\%, CallAtRumors outperforms ML-GRU by 5\% on precision and 4\% on recall on both Twitter and Weibo datasets. The result shows that attention mechanism is more effective in early stage detection by focusing on the most distinct features in advance. With more data applied into test, all methods are approaching their best performance. For Twitter dataset and Weibo Dataset with averagely 80\% duplicate contents in each event, our method starts with 74.02\% and 71.73\% in precision while 68.75\% and 70.34\% in recall, which means an average time lag of 20.47 hours after the emerge of one event. This result is promising because the average report time over the rumors given by Snopes and Sina Community Management Center is 54 hours and 72 hours respectively \cite{ma2016detecting}, and we can save much manual effort with the help of our deep attention based early rumor detection technique.

\section{Conclusion}\label{sec:con}
Rumor detection on social media is time-sensitive because it is hard to eliminate the vicious impact in its late period of diffusion as rumors can spread quickly and broadly. In this paper, we introduce CallAtRumors, a novel recurrent neural network model based on soft attention mechanism to automatically carry out early rumor detection by learning latent representations from the sequential social posts. We conducted experiments with five state-of-the-art rumor detection methods to illustrate that CallAtRumors is sensitive to distinguishable words, thus outperforming the competitors even when textual feature is sparse at the beginning stage of a rumor. In addition, we demonstrate the capability of our model to handle duplicate data with a further comparison. In our future work, it would be appealing to investigate the possibility to combine more complexed feature with our deep attention model. For example, we can model the propagation pattern of rumors as sequential inputs for RNNs to improve the detection accuracy. The future work may investigate the efficiency issue using hashing techniques \cite{YXLSIGIR15,LYIVC} over multi-level feature spaces \cite{LYPTCYB,YXLTIP15,LXCVPR10,YXQLPAKDD14,LYJMM13}.  

\bibliographystyle{ACM-Reference-Format}
\bibliography{sample-sigconf}


\begin{thebibliography}{00}

%%% the Web address.

\ifx \showCODEN    \undefined \def \showCODEN     #1{\unskip}     \fi
\ifx \showDOI      \undefined \def \showDOI       #1{{\tt DOI:}\penalty0{#1}\ }
  \fi
\ifx \showISBNx    \undefined \def \showISBNx     #1{\unskip}     \fi
\ifx \showISBNxiii \undefined \def \showISBNxiii  #1{\unskip}     \fi
\ifx \showISSN     \undefined \def \showISSN      #1{\unskip}     \fi
\ifx \showLCCN     \undefined \def \showLCCN      #1{\unskip}     \fi
\ifx \shownote     \undefined \def \shownote      #1{#1}          \fi
\ifx \showarticletitle \undefined \def \showarticletitle #1{#1}   \fi
\ifx \showURL      \undefined \def \showURL       #1{#1}          \fi
% invisible to TeX
\providecommand\bibfield[2]{#2}
\providecommand\bibinfo[2]{#2}
\providecommand\natexlab[1]{#1}
\providecommand\showeprint[2][]{arXiv:#2}

\bibitem[\protect\citeauthoryear{Ba, Grosse, Salakhutdinov, and Frey}{Ba
  et~al\mbox{.}}{2015}]%
        {Wake-sleep}
\bibfield{author}{\bibinfo{person}{Jimmy Ba}, \bibinfo{person}{Roger Grosse},
  \bibinfo{person}{Ruslan Salakhutdinov}, {and} \bibinfo{person}{Brendan
  Frey}.} \bibinfo{year}{2015}\natexlab{}.
\newblock \showarticletitle{Learning wake-sleep recurrent attention models}. In
  \bibinfo{booktitle}{{\em NIPS}}.
\newblock


\bibitem[\protect\citeauthoryear{Bahdanau, Cho, and Bengio}{Bahdanau
  et~al\mbox{.}}{2014}]%
        {bahdanau2014neural}
\bibfield{author}{\bibinfo{person}{Dzmitry Bahdanau},
  \bibinfo{person}{Kyunghyun Cho}, {and} \bibinfo{person}{Yoshua Bengio}.}
  \bibinfo{year}{2014}\natexlab{}.
\newblock \showarticletitle{Neural machine translation by jointly learning to
  align and translate}.
\newblock \bibinfo{journal}{{\em arXiv preprint arXiv:1409.0473\/}}
  (\bibinfo{year}{2014}).
\newblock


\bibitem[\protect\citeauthoryear{Bengio, Simard, and Frasconi}{Bengio
  et~al\mbox{.}}{1994}]%
        {DifficultRNN1994}
\bibfield{author}{\bibinfo{person}{Yoshua Bengio}, \bibinfo{person}{Patrice
  Simard}, {and} \bibinfo{person}{Paolo Frasconi}.}
  \bibinfo{year}{1994}\natexlab{}.
\newblock \showarticletitle{Learning long-term dependencies with gradient
  decent is difficult}.
\newblock \bibinfo{journal}{{\em IEEE Transactions on Neural Networks\/}}
  \bibinfo{volume}{5}, \bibinfo{number}{2} (\bibinfo{year}{1994}),
  \bibinfo{pages}{157--166}.
\newblock


\bibitem[\protect\citeauthoryear{Castillo, Mendoza, and Poblete}{Castillo
  et~al\mbox{.}}{2011}]%
        {castillo2011information}
\bibfield{author}{\bibinfo{person}{Carlos Castillo}, \bibinfo{person}{Marcelo
  Mendoza}, {and} \bibinfo{person}{Barbara Poblete}.}
  \bibinfo{year}{2011}\natexlab{}.
\newblock \showarticletitle{Information credibility on twitter}. In
  \bibinfo{booktitle}{{\em Proceedings of the 20th international conference on
  World wide web}}. ACM, \bibinfo{pages}{675--684}.
\newblock


\bibitem[\protect\citeauthoryear{Collobert, Weston, Bottou, Karlen,
  Kavukcuoglu, and Kuksa}{Collobert et~al\mbox{.}}{2011}]%
        {collobert2011natural}
\bibfield{author}{\bibinfo{person}{Ronan Collobert}, \bibinfo{person}{Jason
  Weston}, \bibinfo{person}{L{\'e}on Bottou}, \bibinfo{person}{Michael Karlen},
  \bibinfo{person}{Koray Kavukcuoglu}, {and} \bibinfo{person}{Pavel Kuksa}.}
  \bibinfo{year}{2011}\natexlab{}.
\newblock \showarticletitle{Natural language processing (almost) from scratch}.
\newblock \bibinfo{journal}{{\em Journal of Machine Learning Research\/}}
  \bibinfo{volume}{12}, \bibinfo{number}{Aug} (\bibinfo{year}{2011}),
  \bibinfo{pages}{2493--2537}.
\newblock


\bibitem[\protect\citeauthoryear{Goodfellow, Bengio, and Courville}{Goodfellow
  et~al\mbox{.}}{2016}]%
        {Goodfellow-et-al-2016}
\bibfield{author}{\bibinfo{person}{Ian Goodfellow}, \bibinfo{person}{Yoshua
  Bengio}, {and} \bibinfo{person}{Aaron Courville}.}
  \bibinfo{year}{2016}\natexlab{}.
\newblock \bibinfo{booktitle}{{\em Deep Learning}}.
\newblock \bibinfo{publisher}{MIT Press}.
\newblock
\newblock
\shownote{\url{http://www.deeplearningbook.org}.}


\bibitem[\protect\citeauthoryear{Graves}{Graves}{2013}]%
        {graves2013generating}
\bibfield{author}{\bibinfo{person}{Alex Graves}.}
  \bibinfo{year}{2013}\natexlab{}.
\newblock \showarticletitle{Generating sequences with recurrent neural
  networks}.
\newblock \bibinfo{journal}{{\em arXiv preprint arXiv:1308.0850\/}}
  (\bibinfo{year}{2013}).
\newblock


\bibitem[\protect\citeauthoryear{Greff, Srivastava, Koutn{\'\i}k, Steunebrink,
  and Schmidhuber}{Greff et~al\mbox{.}}{2016}]%
        {greff2016lstm}
\bibfield{author}{\bibinfo{person}{Klaus Greff}, \bibinfo{person}{Rupesh~K
  Srivastava}, \bibinfo{person}{Jan Koutn{\'\i}k}, \bibinfo{person}{Bas~R
  Steunebrink}, {and} \bibinfo{person}{J{\"u}rgen Schmidhuber}.}
  \bibinfo{year}{2016}\natexlab{}.
\newblock \showarticletitle{LSTM: A search space odyssey}.
\newblock \bibinfo{journal}{{\em IEEE transactions on neural networks and
  learning systems\/}} (\bibinfo{year}{2016}).
\newblock


\bibitem[\protect\citeauthoryear{Hochreiter and Schmidhuber}{Hochreiter and
  Schmidhuber}{1997}]%
        {lstm1997}
\bibfield{author}{\bibinfo{person}{Sepp Hochreiter} {and}
  \bibinfo{person}{Jurgen Schmidhuber}.} \bibinfo{year}{1997}\natexlab{}.
\newblock \showarticletitle{Long short-term memory}. In
  \bibinfo{booktitle}{{\em Neural computation}}.
\newblock


\bibitem[\protect\citeauthoryear{Hu, Tang, Gao, and Liu}{Hu
  et~al\mbox{.}}{2014}]%
        {hu2014social}
\bibfield{author}{\bibinfo{person}{Xia Hu}, \bibinfo{person}{Jiliang Tang},
  \bibinfo{person}{Huiji Gao}, {and} \bibinfo{person}{Huan Liu}.}
  \bibinfo{year}{2014}\natexlab{}.
\newblock \showarticletitle{Social spammer detection with sentiment
  information}. In \bibinfo{booktitle}{{\em Data Mining (ICDM), 2014 IEEE
  International Conference on}}. IEEE, \bibinfo{pages}{180--189}.
\newblock


\bibitem[\protect\citeauthoryear{Kingma and Ba}{Kingma and Ba}{2014}]%
        {kingma2014adam}
\bibfield{author}{\bibinfo{person}{Diederik Kingma} {and}
  \bibinfo{person}{Jimmy Ba}.} \bibinfo{year}{2014}\natexlab{}.
\newblock \showarticletitle{Adam: A method for stochastic optimization}.
\newblock \bibinfo{journal}{{\em arXiv preprint arXiv:1412.6980\/}}
  (\bibinfo{year}{2014}).
\newblock


\bibitem[\protect\citeauthoryear{Kwon, Cha, Jung, Chen, and Wang}{Kwon
  et~al\mbox{.}}{2013}]%
        {kwon2013prominent}
\bibfield{author}{\bibinfo{person}{Sejeong Kwon}, \bibinfo{person}{Meeyoung
  Cha}, \bibinfo{person}{Kyomin Jung}, \bibinfo{person}{Wei Chen}, {and}
  \bibinfo{person}{Yajun Wang}.} \bibinfo{year}{2013}\natexlab{}.
\newblock \showarticletitle{Prominent features of rumor propagation in online
  social media}. In \bibinfo{booktitle}{{\em Data Mining (ICDM), 2013 IEEE 13th
  International Conference on}}. IEEE, \bibinfo{pages}{1103--1108}.
\newblock


\bibitem[\protect\citeauthoryear{LeCun, Bengio, and Hinton}{LeCun
  et~al\mbox{.}}{2015}]%
        {lecun2015deep}
\bibfield{author}{\bibinfo{person}{Yann LeCun}, \bibinfo{person}{Yoshua
  Bengio}, {and} \bibinfo{person}{Geoffrey Hinton}.}
  \bibinfo{year}{2015}\natexlab{}.
\newblock \showarticletitle{Deep learning}.
\newblock \bibinfo{journal}{{\em Nature\/}} \bibinfo{volume}{521},
  \bibinfo{number}{7553} (\bibinfo{year}{2015}), \bibinfo{pages}{436--444}.
\newblock


\bibitem[\protect\citeauthoryear{Leskovec, Rajaraman, and Ullman}{Leskovec
  et~al\mbox{.}}{2014}]%
        {leskovec2014mining}
\bibfield{author}{\bibinfo{person}{Jure Leskovec}, \bibinfo{person}{Anand
  Rajaraman}, {and} \bibinfo{person}{Jeffrey~David Ullman}.}
  \bibinfo{year}{2014}\natexlab{}.
\newblock \bibinfo{booktitle}{{\em Mining of massive datasets}}.
\newblock \bibinfo{publisher}{Cambridge University Press}.
\newblock


\bibitem[\protect\citeauthoryear{Liu, Huang, Cheng, Chen, Shen, and Zhang}{Liu
  et~al\mbox{.}}{2013}]%
        {Duplicate-photo}
\bibfield{author}{\bibinfo{person}{Jiajun Liu}, \bibinfo{person}{Zi Huang},
  \bibinfo{person}{Hong Cheng}, \bibinfo{person}{Yueguo Chen},
  \bibinfo{person}{Heng~Tao Shen}, {and} \bibinfo{person}{Yanchun Zhang}.}
  \bibinfo{year}{2013}\natexlab{}.
\newblock \showarticletitle{Presenting diverse location views with real-time
  near-duplicate photo elimination}. In \bibinfo{booktitle}{{\em ICDE}}.
\newblock


\bibitem[\protect\citeauthoryear{Liu, Nourbakhsh, Li, Fang, and Shah}{Liu
  et~al\mbox{.}}{2015}]%
        {liu2015real}
\bibfield{author}{\bibinfo{person}{Xiaomo Liu}, \bibinfo{person}{Armineh
  Nourbakhsh}, \bibinfo{person}{Quanzhi Li}, \bibinfo{person}{Rui Fang}, {and}
  \bibinfo{person}{Sameena Shah}.} \bibinfo{year}{2015}\natexlab{}.
\newblock \showarticletitle{Real-time rumor debunking on twitter}. In
  \bibinfo{booktitle}{{\em Proceedings of the 24th ACM International on
  Conference on Information and Knowledge Management}}. ACM,
  \bibinfo{pages}{1867--1870}.
\newblock


\bibitem[\protect\citeauthoryear{Ma, Gao, Mitra, Kwon, Jansen, Wong, and
  Cha}{Ma et~al\mbox{.}}{2016}]%
        {ma2016detecting}
\bibfield{author}{\bibinfo{person}{Jing Ma}, \bibinfo{person}{Wei Gao},
  \bibinfo{person}{Prasenjit Mitra}, \bibinfo{person}{Sejeong Kwon},
  \bibinfo{person}{Bernard~J Jansen}, \bibinfo{person}{Kam-Fai Wong}, {and}
  \bibinfo{person}{Meeyoung Cha}.} \bibinfo{year}{2016}\natexlab{}.
\newblock \showarticletitle{Detecting rumors from microblogs with recurrent
  neural networks}. In \bibinfo{booktitle}{{\em Proceedings of IJCAI}}.
\newblock


\bibitem[\protect\citeauthoryear{Ma, Gao, Wei, Lu, and Wong}{Ma
  et~al\mbox{.}}{2015}]%
        {ma2015detect}
\bibfield{author}{\bibinfo{person}{Jing Ma}, \bibinfo{person}{Wei Gao},
  \bibinfo{person}{Zhongyu Wei}, \bibinfo{person}{Yueming Lu}, {and}
  \bibinfo{person}{Kam-Fai Wong}.} \bibinfo{year}{2015}\natexlab{}.
\newblock \showarticletitle{Detect rumors using time series of social context
  information on microblogging websites}. In \bibinfo{booktitle}{{\em
  Proceedings of the 24th ACM International on Conference on Information and
  Knowledge Management}}. ACM, \bibinfo{pages}{1751--1754}.
\newblock


\bibitem[\protect\citeauthoryear{Mnih, Heess, Graves, and Kavukcuoglu}{Mnih
  et~al\mbox{.}}{2014}]%
        {MnihNIPS2014}
\bibfield{author}{\bibinfo{person}{Volodymyr Mnih}, \bibinfo{person}{Nicolas
  Heess}, \bibinfo{person}{Alex Graves}, {and} \bibinfo{person}{Koray
  Kavukcuoglu}.} \bibinfo{year}{2014}\natexlab{}.
\newblock \showarticletitle{Recurrent models of visual attention}. In
  \bibinfo{booktitle}{{\em NIPS}}.
\newblock


\bibitem[\protect\citeauthoryear{Rayana and Akoglu}{Rayana and Akoglu}{2015}]%
        {rayana2015collective}
\bibfield{author}{\bibinfo{person}{Shebuti Rayana} {and} \bibinfo{person}{Leman
  Akoglu}.} \bibinfo{year}{2015}\natexlab{}.
\newblock \showarticletitle{Collective opinion spam detection: Bridging review
  networks and metadata}. In \bibinfo{booktitle}{{\em Proceedings of the 21th
  ACM SIGKDD International Conference on Knowledge Discovery and Data Mining}}.
  ACM, \bibinfo{pages}{985--994}.
\newblock


\bibitem[\protect\citeauthoryear{Rayana and Akoglu}{Rayana and Akoglu}{2016}]%
        {rayana2016collective}
\bibfield{author}{\bibinfo{person}{Shebuti Rayana} {and} \bibinfo{person}{Leman
  Akoglu}.} \bibinfo{year}{2016}\natexlab{}.
\newblock \showarticletitle{Collective opinion spam detection using active
  inference}. In \bibinfo{booktitle}{{\em Proceedings of the 2016 SIAM
  International Conference on Data Mining}}. SIAM, \bibinfo{pages}{630--638}.
\newblock


\bibitem[\protect\citeauthoryear{Rockt{\"a}schel, Grefenstette, Hermann,
  Ko{\v{c}}isk{\`y}, and Blunsom}{Rockt{\"a}schel et~al\mbox{.}}{2015}]%
        {rocktaschel2015reasoning}
\bibfield{author}{\bibinfo{person}{Tim Rockt{\"a}schel},
  \bibinfo{person}{Edward Grefenstette}, \bibinfo{person}{Karl~Moritz Hermann},
  \bibinfo{person}{Tom{\'a}{\v{s}} Ko{\v{c}}isk{\`y}}, {and}
  \bibinfo{person}{Phil Blunsom}.} \bibinfo{year}{2015}\natexlab{}.
\newblock \showarticletitle{Reasoning about entailment with neural attention}.
\newblock \bibinfo{journal}{{\em arXiv preprint arXiv:1509.06664\/}}
  (\bibinfo{year}{2015}).
\newblock


\bibitem[\protect\citeauthoryear{Rumelhart, Hinton, and Williams}{Rumelhart
  et~al\mbox{.}}{1988}]%
        {rumelhart1988learning}
\bibfield{author}{\bibinfo{person}{David~E Rumelhart},
  \bibinfo{person}{Geoffrey~E Hinton}, {and} \bibinfo{person}{Ronald~J
  Williams}.} \bibinfo{year}{1988}\natexlab{}.
\newblock \showarticletitle{Learning representations by back-propagating
  errors}.
\newblock \bibinfo{journal}{{\em Cognitive modeling\/}} \bibinfo{volume}{5},
  \bibinfo{number}{3} (\bibinfo{year}{1988}), \bibinfo{pages}{1}.
\newblock


\bibitem[\protect\citeauthoryear{Sampson, Morstatter, Wu, and Liu}{Sampson
  et~al\mbox{.}}{2016}]%
        {sampson2016leveraging}
\bibfield{author}{\bibinfo{person}{Justin Sampson}, \bibinfo{person}{Fred
  Morstatter}, \bibinfo{person}{Liang Wu}, {and} \bibinfo{person}{Huan Liu}.}
  \bibinfo{year}{2016}\natexlab{}.
\newblock \showarticletitle{Leveraging the Implicit Structure within Social
  Media for Emergent Rumor Detection}. In \bibinfo{booktitle}{{\em Proceedings
  of the 25th ACM International on Conference on Information and Knowledge
  Management}}. ACM, \bibinfo{pages}{2377--2382}.
\newblock


\bibitem[\protect\citeauthoryear{Sharma, Kiros, and Salakhutdinov}{Sharma
  et~al\mbox{.}}{2015}]%
        {sharma2015action}
\bibfield{author}{\bibinfo{person}{Shikhar Sharma}, \bibinfo{person}{Ryan
  Kiros}, {and} \bibinfo{person}{Ruslan Salakhutdinov}.}
  \bibinfo{year}{2015}\natexlab{}.
\newblock \showarticletitle{Action recognition using visual attention}.
\newblock \bibinfo{journal}{{\em arXiv preprint arXiv:1511.04119\/}}
  (\bibinfo{year}{2015}).
\newblock


\bibitem[\protect\citeauthoryear{Song, Yang, Huang, Shen, and Hong}{Song
  et~al\mbox{.}}{2011}]%
        {Duplicate-video}
\bibfield{author}{\bibinfo{person}{Jingkuan Song}, \bibinfo{person}{Yi Yang},
  \bibinfo{person}{Zi Huang}, \bibinfo{person}{Heng~Tao Shen}, {and}
  \bibinfo{person}{Richang Hong}.} \bibinfo{year}{2011}\natexlab{}.
\newblock \showarticletitle{Multiple feature hashing for real-time large scale
  near-duplicate video retrieval}. In \bibinfo{booktitle}{{\em ACM
  Multimedia}}.
\newblock


\bibitem[\protect\citeauthoryear{Srivastava, Hinton, Krizhevsky, Sutskever, and
  Salakhutdinov}{Srivastava et~al\mbox{.}}{2014}]%
        {srivastava2014dropout}
\bibfield{author}{\bibinfo{person}{Nitish Srivastava},
  \bibinfo{person}{Geoffrey~E Hinton}, \bibinfo{person}{Alex Krizhevsky},
  \bibinfo{person}{Ilya Sutskever}, {and} \bibinfo{person}{Ruslan
  Salakhutdinov}.} \bibinfo{year}{2014}\natexlab{}.
\newblock \showarticletitle{Dropout: a simple way to prevent neural networks
  from overfitting}.
\newblock \bibinfo{journal}{{\em Journal of Machine Learning Research\/}}
  \bibinfo{volume}{15}, \bibinfo{number}{1} (\bibinfo{year}{2014}),
  \bibinfo{pages}{1929--1958}.
\newblock


\bibitem[\protect\citeauthoryear{Sutskever, Vinyals, and Le}{Sutskever
  et~al\mbox{.}}{2014}]%
        {sutskever2014sequence}
\bibfield{author}{\bibinfo{person}{Ilya Sutskever}, \bibinfo{person}{Oriol
  Vinyals}, {and} \bibinfo{person}{Quoc~V Le}.}
  \bibinfo{year}{2014}\natexlab{}.
\newblock \showarticletitle{Sequence to sequence learning with neural
  networks}. In \bibinfo{booktitle}{{\em Advances in neural information
  processing systems}}. \bibinfo{pages}{3104--3112}.
\newblock


\bibitem[\protect\citeauthoryear{Vinyals, Kaiser, Koo, Petrov, Sutskever, and
  Hinton}{Vinyals et~al\mbox{.}}{2015}]%
        {vinyals2015grammar}
\bibfield{author}{\bibinfo{person}{Oriol Vinyals}, \bibinfo{person}{{\L}ukasz
  Kaiser}, \bibinfo{person}{Terry Koo}, \bibinfo{person}{Slav Petrov},
  \bibinfo{person}{Ilya Sutskever}, {and} \bibinfo{person}{Geoffrey Hinton}.}
  \bibinfo{year}{2015}\natexlab{}.
\newblock \showarticletitle{Grammar as a foreign language}. In
  \bibinfo{booktitle}{{\em Advances in Neural Information Processing Systems}}.
  \bibinfo{pages}{2773--2781}.
\newblock


\bibitem[\protect\citeauthoryear{Wang, Lin, Wu, and Zhang}{Wang
  et~al\mbox{.}}{2015}]%
        {YXLMM15}
\bibfield{author}{\bibinfo{person}{Yang Wang}, \bibinfo{person}{Xuemin Lin},
  \bibinfo{person}{Lin Wu}, {and} \bibinfo{person}{Wenjie Zhang}.}
  \bibinfo{year}{2015}\natexlab{}.
\newblock \showarticletitle{Effective Multi-Query Expansions: Robust Landmark
  Retrieval}. In \bibinfo{booktitle}{{\em ACM Multimedia}}.
\newblock


\bibitem[\protect\citeauthoryear{Wang, Lin, Wu, and Zhang}{Wang
  et~al\mbox{.}}{2017}]%
        {YXLTIP17}
\bibfield{author}{\bibinfo{person}{Yang Wang}, \bibinfo{person}{Xuemin Lin},
  \bibinfo{person}{Lin Wu}, {and} \bibinfo{person}{Wenjie Zhang}.}
  \bibinfo{year}{2017}\natexlab{}.
\newblock \showarticletitle{Effective Multi-Query Expansions: Collaborative
  Deep Networks for Robust Landmark Retrieval}.
\newblock \bibinfo{journal}{{\em IEEE Trans. Image Processing\/}}
  \bibinfo{volume}{26}, \bibinfo{number}{3} (\bibinfo{year}{2017}),
  \bibinfo{pages}{1393--1404}.
\newblock


\bibitem[\protect\citeauthoryear{Wang, Lin, Wu, Zhang, and Zhang}{Wang
  et~al\mbox{.}}{2015a}]%
        {YXLSIGIR15}
\bibfield{author}{\bibinfo{person}{Yang Wang}, \bibinfo{person}{Xuemin Lin},
  \bibinfo{person}{Lin Wu}, \bibinfo{person}{Wenjie Zhang}, {and}
  \bibinfo{person}{Qing Zhang}.} \bibinfo{year}{2015}\natexlab{a}.
\newblock \showarticletitle{LBMCH: Learning Bridging Mapping for Cross-modal
  Hashing}. In \bibinfo{booktitle}{{\em ACM SIGIR}}.
\newblock


\bibitem[\protect\citeauthoryear{Wang, Lin, Wu, Zhang, Zhang, and Huang}{Wang
  et~al\mbox{.}}{2015b}]%
        {YXLTIP15}
\bibfield{author}{\bibinfo{person}{Yang Wang}, \bibinfo{person}{Xuemin Lin},
  \bibinfo{person}{Lin Wu}, \bibinfo{person}{Wenjie Zhang},
  \bibinfo{person}{Qing Zhang}, {and} \bibinfo{person}{Xiaodi Huang}.}
  \bibinfo{year}{2015}\natexlab{b}.
\newblock \showarticletitle{Robust Subspace Clustering for Multi-View Data by
  Exploiting Correlation Consensus}.
\newblock \bibinfo{journal}{{\em IEEE Trans. Image Processing\/}}
  \bibinfo{volume}{24}, \bibinfo{number}{11} (\bibinfo{year}{2015}),
  \bibinfo{pages}{3939--3949}.
\newblock


\bibitem[\protect\citeauthoryear{Wang, Lin, and Zhang}{Wang
  et~al\mbox{.}}{2013}]%
        {YXQCIKM13}
\bibfield{author}{\bibinfo{person}{Yang Wang}, \bibinfo{person}{Xuemin Lin},
  {and} \bibinfo{person}{Qing Zhang}.} \bibinfo{year}{2013}\natexlab{}.
\newblock \showarticletitle{Towards metric fusion on multi-view data: a
  cross-view based graph random walk approach}. In \bibinfo{booktitle}{{\em ACM
  CIKM}}.
\newblock


\bibitem[\protect\citeauthoryear{Wang, Lin, Zhang, and Wu}{Wang
  et~al\mbox{.}}{2014}]%
        {YXQLPAKDD14}
\bibfield{author}{\bibinfo{person}{Yang Wang}, \bibinfo{person}{Xuemin Lin},
  \bibinfo{person}{Qing Zhang}, {and} \bibinfo{person}{Lin Wu}.}
  \bibinfo{year}{2014}\natexlab{}.
\newblock \showarticletitle{Shifting Hypergraphs by Probabilistic Voting}. In
  \bibinfo{booktitle}{{\em PAKDD}}.
\newblock


\bibitem[\protect\citeauthoryear{Wang, Zhang, Wu, Lin, and Zhao}{Wang
  et~al\mbox{.}}{2017}]%
        {YLXTNNLS17}
\bibfield{author}{\bibinfo{person}{Yang Wang}, \bibinfo{person}{Wenjie Zhang},
  \bibinfo{person}{Lin Wu}, \bibinfo{person}{Xuemin Lin}, {and}
  \bibinfo{person}{Xiang Zhao}.} \bibinfo{year}{2017}\natexlab{}.
\newblock \showarticletitle{Unsupervised Metric Fusion Over Multiview Data by
  Graph Random Walk-Based Cross-View Diffusion}.
\newblock \bibinfo{journal}{{\em IEEE Trans. Neural Netw. Learning Syst\/}}
  \bibinfo{volume}{28}, \bibinfo{number}{1} (\bibinfo{year}{2017}),
  \bibinfo{pages}{57--70}.
\newblock


\bibitem[\protect\citeauthoryear{Wu, Yang, and Zhu}{Wu et~al\mbox{.}}{2015}]%
        {wu2015false}
\bibfield{author}{\bibinfo{person}{Ke Wu}, \bibinfo{person}{Song Yang}, {and}
  \bibinfo{person}{Kenny~Q Zhu}.} \bibinfo{year}{2015}\natexlab{}.
\newblock \showarticletitle{False rumors detection on sina weibo by propagation
  structures}. In \bibinfo{booktitle}{{\em Data Engineering (ICDE), 2015 IEEE
  31st International Conference on}}. IEEE, \bibinfo{pages}{651--662}.
\newblock


\bibitem[\protect\citeauthoryear{Wu and Cao}{Wu and Cao}{2010}]%
        {LXCVPR10}
\bibfield{author}{\bibinfo{person}{Lin Wu} {and} \bibinfo{person}{Xiaochun
  Cao}.} \bibinfo{year}{2010}\natexlab{}.
\newblock \showarticletitle{Geo-location estimation from two shadow
  trajectories}. In \bibinfo{booktitle}{{\em CVPR}}.
\newblock


\bibitem[\protect\citeauthoryear{Wu, Li, Hu, and Liu}{Wu et~al\mbox{.}}{2016}]%
        {wugleaning}
\bibfield{author}{\bibinfo{person}{Liang Wu}, \bibinfo{person}{Jundong Li},
  \bibinfo{person}{Xia Hu}, {and} \bibinfo{person}{Huan Liu}.}
  \bibinfo{year}{2016}\natexlab{}.
\newblock \showarticletitle{Gleaning Wisdom from the Past: Early Detection of
  Emerging Rumors in Social Media}. In \bibinfo{booktitle}{{\em SDM}}.
\newblock


\bibitem[\protect\citeauthoryear{Wu, Shen, and van~den Hengel}{Wu
  et~al\mbox{.}}{2017}]%
        {LCAPR2017}
\bibfield{author}{\bibinfo{person}{Lin Wu}, \bibinfo{person}{Chunhua Shen},
  {and} \bibinfo{person}{Anton van~den Hengel}.}
  \bibinfo{year}{2017}\natexlab{}.
\newblock \showarticletitle{Deep linear discriminant analysis on fisher
  networks: A hybrid architecture for person re-identification}.
\newblock \bibinfo{journal}{{\em Pattern Recognition\/}}  \bibinfo{volume}{65}
  (\bibinfo{year}{2017}), \bibinfo{pages}{238--250}.
\newblock


\bibitem[\protect\citeauthoryear{Wu and Wang}{Wu and Wang}{2017}]%
        {LYIVC}
\bibfield{author}{\bibinfo{person}{Lin Wu} {and} \bibinfo{person}{Yang Wang}.}
  \bibinfo{year}{2017}\natexlab{}.
\newblock \showarticletitle{Robust hashing for multi-view data: Jointly
  learning low-rank kernelized similarity consensus and hash functions}.
\newblock \bibinfo{journal}{{\em Image Vision Comput\/}}  \bibinfo{volume}{57}
  (\bibinfo{year}{2017}), \bibinfo{pages}{58--66}.
\newblock


\bibitem[\protect\citeauthoryear{Wu, Wang, and Pan}{Wu et~al\mbox{.}}{2016}]%
        {LYPTCYB}
\bibfield{author}{\bibinfo{person}{Lin Wu}, \bibinfo{person}{Yang Wang}, {and}
  \bibinfo{person}{Shirui Pan}.} \bibinfo{year}{2016}\natexlab{}.
\newblock \showarticletitle{Exploiting Attribute Correlations: A Novel Trace
  Lasso based Weakly Supervised Dictionary Learning Method}.
\newblock \bibinfo{journal}{{\em IEEE Trans. Cybernetics\/}}
  (\bibinfo{year}{2016}).
\newblock


\bibitem[\protect\citeauthoryear{Wu, Wang, and Shepherd}{Wu
  et~al\mbox{.}}{2013}]%
        {LYJMM13}
\bibfield{author}{\bibinfo{person}{Lin Wu}, \bibinfo{person}{Yang Wang}, {and}
  \bibinfo{person}{John Shepherd}.} \bibinfo{year}{2013}\natexlab{}.
\newblock \showarticletitle{Efficient image and tag co-ranking: a bregman
  divergence optimization method}. In \bibinfo{booktitle}{{\em ACM
  Multimedia}}.
\newblock


\bibitem[\protect\citeauthoryear{Xiang, Chen, Wang, and Qin}{Xiang
  et~al\mbox{.}}{2017}]%
        {xiang2017answer}
\bibfield{author}{\bibinfo{person}{Yang Xiang}, \bibinfo{person}{Qingcai Chen},
  \bibinfo{person}{Xiaolong Wang}, {and} \bibinfo{person}{Yang Qin}.}
  \bibinfo{year}{2017}\natexlab{}.
\newblock \showarticletitle{Answer Selection in Community Question Answering
  via Attentive Neural Networks}.
\newblock \bibinfo{journal}{{\em IEEE Signal Processing Letters\/}}
  \bibinfo{volume}{24}, \bibinfo{number}{4} (\bibinfo{year}{2017}),
  \bibinfo{pages}{505--509}.
\newblock


\bibitem[\protect\citeauthoryear{Xu, Ba, Kiros, Cho, Courville, Salakhutdinov,
  Zemel, and Bengio}{Xu et~al\mbox{.}}{2015}]%
        {ShowAttendTell}
\bibfield{author}{\bibinfo{person}{Kelvin Xu}, \bibinfo{person}{Jimmy Ba},
  \bibinfo{person}{Ryan Kiros}, \bibinfo{person}{Kyunghyun Cho},
  \bibinfo{person}{Aaron Courville}, \bibinfo{person}{Ruslan Salakhutdinov},
  \bibinfo{person}{Richard Zemel}, {and} \bibinfo{person}{Yoshua Bengio}.}
  \bibinfo{year}{2015}\natexlab{}.
\newblock \showarticletitle{Show, Attend and Tell: Neural Image Caption
  Generation with Visual Attention}. In \bibinfo{booktitle}{{\em ICML}}.
\newblock


\bibitem[\protect\citeauthoryear{Yang, Yang, Dyer, He, Smola, and Hovy}{Yang
  et~al\mbox{.}}{2016}]%
        {yang2016hierarchical}
\bibfield{author}{\bibinfo{person}{Zichao Yang}, \bibinfo{person}{Diyi Yang},
  \bibinfo{person}{Chris Dyer}, \bibinfo{person}{Xiaodong He},
  \bibinfo{person}{Alex Smola}, {and} \bibinfo{person}{Eduard Hovy}.}
  \bibinfo{year}{2016}\natexlab{}.
\newblock \showarticletitle{Hierarchical attention networks for document
  classification}. In \bibinfo{booktitle}{{\em Proceedings of NAACL-HLT}}.
  \bibinfo{pages}{1480--1489}.
\newblock


\bibitem[\protect\citeauthoryear{Zafarani and Liu}{Zafarani and Liu}{2015}]%
        {zafarani201510}
\bibfield{author}{\bibinfo{person}{Reza Zafarani} {and} \bibinfo{person}{Huan
  Liu}.} \bibinfo{year}{2015}\natexlab{}.
\newblock \showarticletitle{10 bits of surprise: Detecting malicious users with
  minimum information}. In \bibinfo{booktitle}{{\em Proceedings of the 24th ACM
  International on Conference on Information and Knowledge Management}}. ACM,
  \bibinfo{pages}{423--431}.
\newblock


\bibitem[\protect\citeauthoryear{Zaremba, Sutskever, and Vinyals}{Zaremba
  et~al\mbox{.}}{2014}]%
        {zaremba2014recurrent}
\bibfield{author}{\bibinfo{person}{Wojciech Zaremba}, \bibinfo{person}{Ilya
  Sutskever}, {and} \bibinfo{person}{Oriol Vinyals}.}
  \bibinfo{year}{2014}\natexlab{}.
\newblock \showarticletitle{Recurrent neural network regularization}.
\newblock \bibinfo{journal}{{\em arXiv preprint arXiv:1409.2329\/}}
  (\bibinfo{year}{2014}).
\newblock


\bibitem[\protect\citeauthoryear{Zhao, Resnick, and Mei}{Zhao
  et~al\mbox{.}}{2015}]%
        {zhao2015enquiring}
\bibfield{author}{\bibinfo{person}{Zhe Zhao}, \bibinfo{person}{Paul Resnick},
  {and} \bibinfo{person}{Qiaozhu Mei}.} \bibinfo{year}{2015}\natexlab{}.
\newblock \showarticletitle{Enquiring minds: Early detection of rumors in
  social media from enquiry posts}. In \bibinfo{booktitle}{{\em Proceedings of
  the 24th International Conference on World Wide Web}}. ACM,
  \bibinfo{pages}{1395--1405}.
\newblock


\bibitem[\protect\citeauthoryear{Zhou, Wang, Wu, Yang, and Sun}{Zhou
  et~al\mbox{.}}{2017}]%
        {ImgCopy}
\bibfield{author}{\bibinfo{person}{Zhili Zhou}, \bibinfo{person}{Yunlong Wang},
  \bibinfo{person}{Q.~M.~Jonathan Wu}, \bibinfo{person}{Ching-Nung Yang}, {and}
  \bibinfo{person}{Xingming Sun}.} \bibinfo{year}{2017}\natexlab{}.
\newblock \showarticletitle{Effective and Efficient Global Context Verification
  for Image Copy Detection}.
\newblock \bibinfo{journal}{{\em IEEE Transactions on Information Forensics and
  Security\/}} \bibinfo{volume}{12}, \bibinfo{number}{1}
  (\bibinfo{year}{2017}), \bibinfo{pages}{48--63}.
\newblock


\bibitem[\protect\citeauthoryear{Zimbra, Ghiassi, and Lee}{Zimbra
  et~al\mbox{.}}{2016}]%
        {zimbra2016brand}
\bibfield{author}{\bibinfo{person}{David Zimbra}, \bibinfo{person}{M Ghiassi},
  {and} \bibinfo{person}{Sean Lee}.} \bibinfo{year}{2016}\natexlab{}.
\newblock \showarticletitle{Brand-Related Twitter Sentiment Analysis Using
  Feature Engineering and the Dynamic Architecture for Artificial Neural
  Networks}. In \bibinfo{booktitle}{{\em System Sciences (HICSS), 2016 49th
  Hawaii International Conference on}}. IEEE, \bibinfo{pages}{1930--1938}.
\newblock


\end{thebibliography}

\end{document}